%% file: CorrAttention_arxiv.tex
\documentclass{article}

\usepackage{arxiv}

\usepackage{natbib}

\usepackage{algorithm}
\usepackage{algorithmic}

\usepackage{hyperref}
\usepackage{url}

\usepackage{multirow}

\title{Correlated Attention in Transformers for Multivariate Time Series}

\author{%
 Quang Minh Nguyen\thanks{This work was done while the author was an intern at MIT-IBM Watson AI Lab.}\\
Electrical Engineering and Computer Science\\
  Massachusetts Institute of Technology\\
 Cambridge, MA, USA \\
  \texttt{nmquang@mit.edu} \\
   \And
   Lam M. Nguyen \\
    IBM Research \\
   Thomas J. Watson Research Center\\
   Yorktown Heights, NY, USA\\
   \texttt{LamNguyen.MLTD@ibm.com} \\
   \And
    Subhro Das \\
    MIT-IBM Watson AI Lab\\
    IBM Research \\
    Cambridge, MA, USA \\
   \texttt{subhro.das@ibm.com} \\
}

\input{import}

\begin{document}

\maketitle

\begin{abstract}

Multivariate time series (MTS)  analysis prevails in real-world applications such as finance, climate science and healthcare. The various self-attention mechanisms, the backbone of the state-of-the-art Transformer-based models, efficiently discover the temporal dependencies, yet cannot well capture the intricate cross-correlation between different features of MTS data, which inherently stems from complex dynamical systems in practice. To this end, we propose a novel correlated attention mechanism, which not only efficiently captures feature-wise dependencies, but can also be seamlessly integrated within the encoder blocks of existing well-known Transformers to gain efficiency improvement. In particular, correlated attention operates across feature channels to compute cross-covariance matrices between queries and keys with different lag values, and selectively aggregate representations at the sub-series level. This architecture facilitates automated discovery and representation learning of not only instantaneous but also lagged cross-correlations, while inherently capturing time series auto-correlation. When combined with prevalent Transformer baselines, correlated attention mechanism constitutes a better alternative for encoder-only architectures, which are suitable for a wide range of tasks including  imputation, anomaly detection and classification. Extensive experiments on the aforementioned tasks consistently underscore the advantages of correlated attention mechanism in enhancing base Transformer models, and demonstrate our state-of-the-art results in imputation, anomaly detection and classification.

\end{abstract}

\input{sections/introduction}

\input{sections/related_work}

\input{sections/methodology}

\input{sections/experiments}

\input{sections/conclusion}

\clearpage


\appendix


\section*{APPENDIX}

\input{sections/appendix}

\bibliographystyle{plainnat}


\bibliography{all_refs,reference}





\end{document}

%% file: import.tex
\usepackage{graphicx}
\usepackage{booktabs} 

\usepackage[flushleft]{threeparttable}

\usepackage{epsf}
\usepackage{fancyhdr}
\usepackage{graphics}
\usepackage{graphicx}
\usepackage{psfrag}
\usepackage{microtype}

\usepackage{color}
\usepackage{amsfonts}
\usepackage{amsmath}
\usepackage{amssymb,bbm}

\usepackage{algorithm}
\usepackage{algorithmic}

\usepackage{caption,subcaption}
\usepackage{sidecap}
\sidecaptionvpos{figure}{c}

\usepackage{url}
\usepackage{hyperref}

\long\def\comment#1{}
\comment{
\setlength{\topmargin}{0 in}
\setlength{\textwidth}{5.5 in}
\setlength{\textheight}{8 in}
\setlength{\oddsidemargin}{0.5 in}
}

\newtheorem{theorem}{Theorem}[section]

\newtheorem{assumption}[theorem]{Assumption}

\newcommand{\Br}{\mathbb{R}}

\def\Ff{\mathcal{F}}

\def\Ss{\mathcal{S}}

\def\Ff{\mathcal{F}}

\def\Xx{\mathcal{X}}

\def\Yy{\mathcal{Y}}

\def\qQ{\mathbf{q}}
\def\kK{\mathbf{k}}

\graphicspath{{./image/}}

\newcommand{\ba}{\begin{array}}
\newcommand{\ea}{\end{array}}

\newcommand{\ones}{\mathbf{1}}

\newcommand{\argtopk}{\mathop{\mathrm{argTopK}}}

\usepackage{psfrag}

\usepackage{amsfonts}
\usepackage{booktabs}
\usepackage{siunitx}

\usepackage{array}
\usepackage{makecell}

%% file: sections/introduction.tex
\section{Introduction}

Multivariate time series (MTS)  are time series  encompassing multiple dimensions for capturing different features of the original data, where each dimension corresponds to a univariate  time series. MTS analysis is ubiquitous in real-world applications such as imputation of missing data in geoscience \citep{LOPEZ2021104794}, anomaly detection of monitoring data in aeronautics \citep{spacecraft1},  classification of hearbeat data for  fetal assessment \citep{heartbeat1}, and weather prediction \citep{wu2022autoformer}. Thanks to its immense practical value, there has been increasing interest in MTS analysis \citep{wen2023transformers, wu2023timesnet, Lim_2021, zhang2023crossformer}.

The recent advancement of deep learning has facilitated the development of many models with superior performance \citep{Li2021SynergeticLO, wu2023timesnet}. Specifically, the large class of  Transformer-based models \citep{wen2023transformers, wu2022autoformer, zhang2023crossformer, fedformer1, liu2022nonstationary, NIPS2017_3f5ee243, DU2023119619} is the most prominent and has demonstrated great potential for their well-known capability to model both short-range and long-range temporal dependencies \citep{wen2023transformers}. In addition to temporal dependencies, feature-wise dependencies, which are cross-correlation between the variates of MTS, are  central to MTS analysis \citep{spectral_temporal1} and studied in the deep learning literature via  convolution neural network (CNN) \citep{lai2018modeling}  or graph neural network (GNN) \citep{mts_forecast_gnn, spectral_temporal1}. Nevertheless,  for existing Transformer-based models (e.g. \citep{enhance_memory1, zhou2021informer, wu2022autoformer}), the embedding method is insufficient for capturing such cross-correlation between different variates of MTS \citep{zhang2023crossformer}, which motivated the authors therein to propose CrossFormer as the first Transformer explicitly utilizing feature-wise dependencies for MTS forecasting. Despite its promising performance, CrossFormer deploys a convoluted architecture, which is isolated from other prevalent Transformers with their own established merits in temporal modelling and specifically designed for only  MTS forecasting, thereby lacking flexibility. 
Consequently, it  remains under-explored whether modelling feature-wise dependencies    could also improve Transformer-based models' performances in other non-predictive tasks, which cover a wide range of real-world applications and include prominently imputation, anomaly detection and classification.  Moreover,  all the previous work \citep{mts_forecast_gnn, spectral_temporal1, zhang2023crossformer} on capturing feature-wise dependencies in MTS analysis are limited in scope to forecasting, rely on  ad-hoc mechanisms in their rigid pipelines, and thus do not fully leverage the  capability to model temporal dependencies of existing powerful Transformers. Motivated by the nascent literature of the aforementioned problems and the success of Transformer-based models in MTS analysis, we  raise the following central question of this paper:  
\begin{center}
    \emph{How can we seamlessly elevate the broad class of existing and future Transformer-based architectures to also capture feature-wise dependencies? Can modelling feature-wise dependencies  improve Transformers' performance on non-predictive tasks?}
\end{center}
We affirmatively answer this question by proposing  a novel correlated attention mechanism that efficiently learns  the cross-correlation  between different variates of MTS and can be seamlessly integrated with the encoder-only architecture of well-known Transformers, thereby being applicable to a wide range of non-predictive tasks. In addition to the conventional cross-correlation, the correlated attention  captures simultaneously auto-correlation, the backbone of Autoformer \citep{wu2022autoformer}, and lagged cross-correlation. Lagged cross-correlation has been inherently critical in MTS data \citep{John2021-mr, Chandereng2020-zz}, yet vastly ignored by the literature of Transformer-based models. For  raw MTS data of production planning (e.g. \citep{CONTRERASREYES2020125109}) as an example, it may take some lagged interval for the increase in the demand rate to be reflected in the production rate. 
Instead of the usual temporal dimension, correlated attention operates across feature channels to compute cross-covariance matrices of between queries and keys with different lag values, and further select the pairs with highest correlations for  aggregating representations at the sub-series level. For seamless integration with  the encoder block of base Tranformers such as \citep{NIPS2017_3f5ee243, liu2022nonstationary} with their respective temporal attentions, the original multi-head attention is modified to include the heads using both the temporal attentions from the base model and our correlated attentions. This design directly augments the embedded layer of the base Transformer with cross-correlation information  in its representation learning. Experimentally, correlated attention, when plugged into prevalent Transformer baselines, consistently boosts the performance of the base models and results in state-of-the-art benchmark for Transformer-models in various tasks . The contributions of the paper can be summarized as follows: 



\begin{itemize}
    \item We propose a novel correlated attention mechanism that efficiently learns both the instantaneous and lagged cross-correlations between different variates of MTS, as well as auto-correlation of series. To the best of our knowledge, this is the first work that presents a Transformer architecture that aims to explicitly learn the lagged cross-correlation. 
    \item Correlated attention is flexible and efficient, where it can be can be seamlessly plugged into  encoder-only architectures of well-known Transformers  such as \citep{NIPS2017_3f5ee243, liu2022nonstationary} to enhance the performance of the base models. It naturally augments the embedded layer of base Transformers, having been known vastly for temporal modelling \citep{zhang2023crossformer}, with feature-wise dependencies.
    Furthermore, the modularity  of correlated attention will permit its adoption in and benefit future Transformer architectures.
    \item  Extensive experiments on imputation, anomaly detection and classification demonstrate that correlated attention consistently improves the performance of base Transformers and results state-of-the-art architectures for the aforementioned tasks. 
\end{itemize}





%% file: sections/related_work.tex
\section{Related Work}

\textbf{Multivariate Time Series Analysis.} The surge of advanced sensors and data stream infrastructures has lead to the tremendous proliferation of MTS data \citep{lit1, lit2}. In response, MTS analysis, which spans a multitude of tasks including but not limiting to imputation \citep{DU2023119619}, anomaly detection \citep{blázquezgarcía2020review}, classification \citep{Ismail_Fawaz_2019} and forecasting \citep{Lim_2021}, has been increasingly crucial. In recent years, many deep learning models have been proposed for MTS analysis and achieved competitive performance \citep{lai2018modeling, franceschi2020unsupervised, wen2023transformers, gu2022efficiently}. Specifically, multilayer perceptron (MLP) methods \citep{oreshkin2020nbeats, challu2022nhits} adopt MLP blocks for modelling temporal dependencies. Temporal Convolutional Networks (TCNs) \citep{lea2016temporal, franceschi2020unsupervised} leverage CNN or recurrent neural network (RNN) along the temporal dimension to capture temporal dependencies. RNN-based models \citep{lstm1, lai2018modeling} use  state transitions and recurrent structure to model temporal variations. In order to capture cross-correlation, recent work \citep{Yu_2018, spectral_temporal1, mts_forecast_gnn}  deploy GNNs to directly model cross-dimension dependencies. Nevertheless, these neural networks rely on RNN and CNN to model temporal dynamics, which are are known to be inefficient in capturing long-range temporal dependencies \citep{zhang2023crossformer}. TimesNet \citep{wu2023timesnet} models temporal 2D-variation for both intraperiod and interperiod variations via residual structure TimesBlock.

\textbf{Transformers in MTS Analysis.} Originating from  natural language processing (NLP) domain, Transformers \citep{NIPS2017_3f5ee243} have shown great success when adapted to MTS analysis \citep{fedformer1, enhance_memory1, zhou2021informer, liu2022nonstationary, wu2022autoformer, DU2023119619} thanks to their capability to capture both short-range and long-range  temporal dependencies \citep{wen2023transformers}. Recently, \cite{liu2022nonstationary} performed series stationarization  to attenuate time series non-stationarity. \cite{wu2022autoformer} proposed  Autoformer with decomposition architecture and auto-correlation mechanism for better modelling of long-range temporal dependencies. Crossformer \citep{zhang2023crossformer} uses dimension-segment-wise embedding and a hierarchical architecture to better learn both the cross-time and cross-dimension dependencies. 
  
\textbf{Modelling Cross-correlation in Time Series.}  Capturing feature-wise dependencies in MTS analysis has been a long lasting problem, where such cross-correlation in MTS data  stems from  natural processes \citep{correlatedMTS2} and complex cyper-physical systems (CPSs) \citep{correlatedMTS1, cirstea2018correlated}. Accurate forecasting of correlated MTS can reveal the underlying  dynamics of the system including trend and intrinsic behavior \citep{correlatedMTS3}, and detect outliers  \citep{correlatedMTS4}. To capture the MTS correlation, previous work have proposed the adoptions of hidden Markov models \citep{correlatedMTS5} and  spatio-temporal (ST) graphs \citep{correlatedMTS6} as the modeling primitives, specialized neural network architectures for correlated MTS forecasting \citep{correlatedMTS1, cirstea2018correlated}, and methods based on cross-correlation analysis \citep{Yuan2016, KRISTOUFEK2014291}. Nevertheless, most of these approaches focused on either forecasting with ST correlation, which  arises from the proximity of the MTS sensors' locations and is only applicable to CPSs, or ad-hoc MTS analysis.  \citep{lai2018modeling} models long and short term temporal patterns with deep neural networks in MTS forecasting. 
Crossformer \citep{zhang2023crossformer} was the first Transformer-based architecture that explicitly utilizes both temporal and feature-wise dependencies for MTS forecasting. Yet, for non-predictive tasks such as imputation, anomaly detection and classification, there has been no Transformer with specialized modelling of feature-wise dependencies. Moreover, while lagged cross-correlation is inherent in MTS data, for which various statistical tools \citep{John2021-mr, Chandereng2020-zz, lagged_cross1, SHEN2015680} have been developed for testing and analysis, time series Transformers in the literature have not leveraged this information in their mechanisms to improve performance of target applications.

%% file: sections/methodology.tex
\section{Methodology}

In this Section, we first review the  two representative well-known temporal attention mechanisms, namely the self-attention  \citep{NIPS2017_3f5ee243} and de-stationary attention \citep{liu2022nonstationary}, and the multi-head attention architecture commonly used in a wide range of Transformer-based models such as \citep{NIPS2017_3f5ee243, liu2022nonstationary, Du_2023, zhou2021informer, wu2022autoformer} and more. 
Next, we discuss the current limitation of conventional temporal attentions in modelling feature-wise dependencies. This then motivates us to  propose the correlated attention mechanism, which operates across the feature channels for learning cross-correlation among variates, and combine it with existing temporal attentions in the mixture-of-head attention architecture to improve the performance of the base Transformers. 

\subsection{Background}
\label{sec_background}

\textbf{Self-attention.} Self-attention, first proposed in the vanilla Transformer \citep{NIPS2017_3f5ee243}, operates on the query, key and value matrices. In particular, given the  input the matrix $X \in \Br^{T \times d}$, where $T$ is the sequence length and $d$ is feature dimension of the model, the model linearly projects $X$ into queries, keys and values respectively as $Q = X W^Q, K = X W^K$ and $V = X W^V$, where $W^Q \in \Br^{d \times d_k}, W^K \in \Br^{d \times d_k}$ and $W^V \in \Br^{d \times d_v}$ are parameter matrices. 
Taking queries $Q$, keys $K$ and values $V$ as input, the self-attention  returns the output  matrix as follows:
\begin{align}
\label{self_attention}
   \textsc{Self-Attention}(Q, K, V) = \textsc{softmax}\bigg(\frac{1}{\sqrt{d_k}} QK^\top\bigg) V.
\end{align}
The computational complexity of self-attention is $O(d_k T^2)$ due to pairwise interactions along the time dimension $T$.


\textbf{De-stationary Attention.} To handle non-stationary real-world MTS data, Non-stationary Transformer 
 \citep{liu2022nonstationary}  performs series stationarization for better predictability and adopts the de-stationary attention mechanism to alleviate the  over-stationarization and recover the intrinsic information into temporal dependencies. Specifically, after the normalization module, Non-stationary Transformer operates over   the stationarized series  $X' = (X - \ones \mu_X^\top)/\sigma_X$ with the mean vector $\mu_X$ and covariance $\sigma_X$, and obtain the stationarized queries, keys and values respectively as $Q' = (K - \ones \mu_Q^\top)/\sigma_X$, $K' = (K - \ones \mu_K^\top)/\sigma_X$ and $V' = (V - \ones \mu_V^\top)/\sigma_X$ with the mean vectors $\mu_Q, \mu_K$ and $\mu_V$. Then, it can be proven that \citep{liu2022nonstationary}:
 \begin{align*}
      \textsc{softmax}\bigg(\frac{1}{\sqrt{d_k}} QK^\top\bigg) =  \textsc{softmax}\bigg(\frac{1}{\sqrt{d_k}} ( \sigma_X^2 Q'K'^\top  + \ones \mu_Q^\top K^\top )\bigg),
 \end{align*}
 which motivates their design of de-stationary attention utilizing  multilayer perceptron (MLP) layer to directly learn the positive scaling scalar $\xi \approx \sigma_X^2$  and shifting vector $\Delta \approx K \mu_Q$, and returning the output matrix: 
 \begin{align}
 \label{destationary_attention}
     \textsc{De-stationary-Attention}(Q', K', V') =  \textsc{softmax}\bigg(\frac{1}{\sqrt{d_k}} ( \xi Q'K'^\top  + \ones \Delta^\top )\bigg) V'.
 \end{align}
 The computational complexity of de-stationary attention is $O(d_k T^2)$ without accounting for the MLP module.

While there have been a multitude of other temporal attention mechanisms (e.g. \citep{zhou2021informer, DU2023119619, fedformer1}) that  usually follow ad-hoc design for specific tasks, the two representative attention mechanisms above are the backbones of some of most primitive Transformers that have robust and competitive performances on a variety of tasks. Next, we present the multi-head attention module, which adopts the temporal attention as its component and commonly used in a wide range of Transformer-based models (e.g. \citep{NIPS2017_3f5ee243, liu2022nonstationary, Du_2023, zhou2021informer}). 
 
\textbf{Multi-head Attention.} Multi-head attention, proposed along with self-attention in the vanilla Transformer \citep{NIPS2017_3f5ee243}, combines multiple temporal attentions to jointly attend to information from different representation subspaces. In particular, it concatenates $h$ heads, where each head is the output from some temporal attention and $h$ is a hyperparameter, and then performs linear projection for the final output. Formally, multi-head attention is written as follows:
\begin{align}
\label{multihead_attention}
\begin{split}
    &\textsc{Multi-head-Attention}(X) = \textsc{Concat}(head_1, head_2, ..., head_h) W^O \\
    \text{wh}&\text{ere } head_i = \textsc{Temporal-Attention}(X W^Q_i, X W^K_i, X W^V_i).
\end{split}
\end{align}
In the Equation \ref{multihead_attention}, $W^O \in \Br^{hd_v \times d}$ is parameter matrix and \textsc{Temporal-Attention} can  take the form of any mechanism, such as the two aforementioned self-attention and de-stationary attention, or any other in the literature \citep{NIPS2017_3f5ee243, liu2022nonstationary, Du_2023, zhou2021informer}.


\subsection{Correlated Attention Block and Mixture-of-head Attention}

In this Section, we first take a deeper look at how the design of self-attention (or more generally temporal attention) can limit its  capability of modeling feature-wise dependencies, while approaches in the literature of Transformers' attention design  may be insufficient to capture the cross-correlation in MTS. This motivates us to propose the correlated attention block (CAB) to efficiently learn the feature-wise dependencies and can be seamlessly plugged into ubiquitous encoder-only Transformer architectures for performance improvement. Next, we demonstrate how the computation for CAB can be further accelerated via  Fast Fourier Transform (FFT) thanks to the Cross-correlation Theorem. 

\subsubsection{Limitation of Temporal Attention}
One interpretation for the powerful temporal modeling capacity of Transformers  is that, with the queries  $Q = [\qQ_1, \qQ_2, ..., \qQ_T]^\top$ and keys $K = [\kK_1, \kK_2, ..., \kK_T]^\top$ expressed in time-wise dimension, the matrix $Q K^\top \in \Br^{T\times T}$ in the computation of self-attention (Equation \ref{self_attention}) contains pairwise inner-products $\qQ_i^\top \kK_j$ of time-dimension vectors, and thus intuitively resembles the notion of correlation matrix between different time points of MTS data. Nevertheless, feature-wise information, where each of the $d_k$ features corresponds to an entry of $\qQ_i \in \Br^{d_k \times 1}$ or $\kK_j \in \Br^{d_k \times 1}$, is absorbed into such  inner-product representation; this thus makes self-attention unable to explicitly leverage the feature-wise information in its representation learning. In the context of computer vision,  \cite{elnouby2021xcit} considered a cross-covariance attention mechanism that instead computes $\hat{K}^\top \hat{Q} \in \Br^{d_k \times d_k} $, where $\hat{K}$ and $\hat{Q}$ are $\ell_2$-normalized versions of $K$ and $Q$, as the cross-covariance matrix along the feature dimension. However, while this simple design is suitable for capturing instantaneous cross-correlation in static image applications as considered therein, it is insufficient to capture the cross-correlation of MTS data which is  coupled with the intrinsic temporal dependencies. In particular, the variates of MTS data can be correlated with each other, yet with a lag interval-- this phenomenon is referred to as lagged cross-correlation in MTS analysis \citep{John2021-mr, Chandereng2020-zz, lagged_cross1, SHEN2015680}. Additionally, a variate in MTS data can even be correlated with the delayed copy of itself, the phenomenon of which is termed auto-correlation. \cite{wu2022autoformer} proposed Autoformer with the auto-correlation mechanism, but their rigid framework is specifically designed for and achieves competitive performance in long-term forecasting. Given the nascent literature of modules to augment a broad class of powerful Transformers with yet less-efficient modelling capabilities of cross-correlation and auto-correlation, we hereby aim to derive a flexible and  efficient correlated attention mechanism that can  elevate existing Transformer-based models.


\subsubsection{Correlated Attention Block}
We proceed to present our correlated attention block (CAB), which is comprised of three consecutive components: normalization (Equation \ref{letstopk1}), lagged cross-correlation filtering (Equation \ref{letstopk2}), and score aggregation (Equation \ref{letstopk3}). 


\textbf{Normalization.} In the normalization step, we perform column-wise $\ell_2$ normalization of $Q$ and $K$, respectively resulting in $\hat{Q}$ and $\hat{K}$ as:
\begin{align}
    \hat{Q} = \textsc{normalize}&(Q), \quad \hat{K} = \textsc{normalize}(K), \label{letstopk1}. 
\end{align}

\textbf{Lagged Cross-correlation Filtering.} We first present  the overview of the lagged cross-correlation filtering step as follows:
\begin{align}
    \begin{split}
    l_1, l_2, ..., l_k &= \argtopk_{l \in [1, T-1]}\bigg\{ \lambda \cdot \textsc{diagonal}\big(\textsc{roll}(\hat{K}, l)^\top \hat{Q}\big) \\
    &\quad  + (1- \lambda) \cdot \textsc{non-diagonal}\big(\textsc{roll}(\hat{K}, l)^\top \hat{Q}\big)  \bigg\}, 
    \end{split}\label{letstopk2}
\end{align}
where $\lambda \in  [0,1]$ is a learnable parameter and $\argtopk(.)$ is used to select the $k = c \lceil \log(T) \rceil$ (with $c$ being a hyperparameter)  time lags  which incur the highest cross-correlation scores to be described in more details now. The purpose of the previous normalization step is  to unify the feature-wise variates into the same scale, so that $\textsc{roll}(\hat{K}, l)^\top \hat{Q}$  can better serve as a notion of cross-correlation matrix in feature-wise dimension between that queries $\hat{Q}$ and the lagged keys $\textsc{roll}(\hat{K}, l)$. Here, for $X\in \Br^{T \times d_k}$, the $\textsc{roll}(X ,  l)$ operation shifts the elements of $X$ vertically, i.e. along the time-dimension, during which entries shifted over the first position are then re-introduced at the last position. This rolling operation helps generating lagged series representation. In order to formally  define our lagged cross-correlation filtering step (Equation \ref{letstopk2}), we hereby  consider the two operations $\textsc{diagonal}(.)$ and $\textsc{non-diagonal}(.)$ on square matrix that respectively sum up the absolute values of diagonal entries and non-diagonal entries. Specifically, given a matrix $A\in \Br^{d_k \times d_k}$, we then have:
\begin{align*}
    \textsc{diagonal}(A) &= \sum_{i=1}^{d_k} | A_{ii}|,\\
    \textsc{non-diagonal}(A) &= \sum_{i, j \in [1, d_k]: i\neq j} | A_{ij}|.
\end{align*}
Recall from  stochastic process theory \citep{chatfield2004timeseries, Papoulis1965ProbabilityRV}  that for any real discrete-time process $\{\Xx_t\}$, its auto-correlation $R_{\Xx,\Xx}(l)$ can be computed by  $R_{\Xx, \Xx}(l) = \lim_{L \to \infty}\frac{1}{L} \sum_{t=1}^L \Xx_t \Xx_{t-l}$. With the normalized queries  $\hat{Q} = [\hat{\qQ}_1, \hat{\qQ}_2, ..., \hat{\qQ}_{d_k}]$ and normalized keys $\hat{K} = [\hat{\kK}_1, \hat{\kK}_2, ..., \hat{\kK}_{d_k}]$ expressed in feature-wise dimension where $\hat{\qQ}_i, \hat{\kK}_j \in \Br^{T \times 1}$, any $i^{th}$ diagonal entry of $\textsc{roll}(\hat{K}, l)^\top \hat{Q}$ takes the form $\big( \textsc{roll}(\hat{K}, l)^\top \hat{Q}\big)_{ii}= R_{ \hat{\qQ}_i, \hat{\kK}_i}(l)= \sum_{t= 1}^T (\hat{\qQ}_i)_t  \cdot (\hat{\kK}_i)_{t-l}$ and thus can serve as an approximation (with multiplicative factor) for the auto-correlation of variate $i$. This idea was   also harnessed in the design of auto-correlation attention \citep{wu2022autoformer}. Consequently, given a lag $l$, the quantity $\textsc{diagonal}\big(\textsc{roll}(\hat{K}, l)^\top \hat{Q}\big)$, which aggregates over the absolute values of all diagonal entries, scores the total auto-correlation of all the feature variates, while the quantity $\textsc{non-diagonal}\big(\textsc{roll}(\hat{K}, l)^\top \hat{Q}\big)$ scores the the total cross-correlation between different pairs of feature variates. The final cross-correlation score incurred by time lag $l$ is then the weighted (convex) combination of  $\textsc{diagonal}\big(\textsc{roll}(\hat{K}, l)^\top \hat{Q}\big)$ and $\textsc{non-diagonal}\big(\textsc{roll}(\hat{K}, l)^\top \hat{Q}\big)$ with a learnable weight $\lambda$ as shown in Equation \ref{letstopk2}. For high-dimensional MTS data where not all pairs of variates are highly correlated and/or auto-correlation is the more significant factor, the learnable parameter $\lambda$ helps automatically untangle such relations and balance the representation learning between auto-correlation and cross-correlation of interacting features. Then $k = c \lceil \log(T) \rceil$ (with $c$ being a hyperparameter)  time lags $l_1, l_2, ..., l_k$, which get the highest cross-correlation scores, are selected through the TopK operation to be used in the next step. 

\textbf{Score Aggregation.} Finally, the CAB performs sub-series aggregation for the final output via:
\begin{align}
    \nonumber
    \textsc{Correlated-Attention}(Q, V, K) &= (1- \beta) \cdot  \textsc{roll}({V}, 0) \cdot  \textsc{softmax}\bigg(\frac{1}{\tau} \textsc{roll}(\hat{K}, 0)^\top \hat{Q} \bigg) \\
    &\quad + \beta \cdot  \sum_{i=1}^k \textsc{roll}({V}, l_i) \cdot \textsc{softmax}\bigg(\frac{1}{\tau} \textsc{roll}(\hat{K}, l_i)^\top \hat{Q} \bigg),   \label{letstopk3}
\end{align}
where $\beta \in [0, 1]$ and $\tau >0$ are  learnable parameters. In particular, for every chosen lag $l_i$, we also roll the values matrix $V$ by $l_i$ to align similar sub-series with the same phase position. Then, each  $ \textsc{roll}({V}, l_i) \cdot \textsc{softmax}\big(\frac{1}{\tau} \textsc{roll}(\hat{K}, l_i)^\top \hat{Q} \big)$ is a convex combination in feature dimension (as opposed to time dimension in self-attention in Equation \ref{self_attention}) of the corresponding token embedding in the delayed values   $ \textsc{roll}({V}, l_i) $. The final score aggregation in Equation \ref{letstopk3} is  the weighted (convex) combination of the ``instantaneous" score $ \textsc{roll}({V}, 0) \cdot  \textsc{softmax}\big(\frac{1}{\tau} \textsc{roll}(\hat{K}, 0)^\top \hat{Q} \big)$ and the  ``lagged" total score $ \sum_{i=1}^k \textsc{roll}({V}, l_i) \cdot \textsc{softmax}\big(\frac{1}{\tau} \textsc{roll}(\hat{K}, l_i)^\top \hat{Q} \big)$ with a learnable weight $\beta$.

\textbf{Efficient computation of CAB.} In its current form, the computation complexity of CAB is $O(d_k^2 T^2)$. Specifically, for every lag $l$, the computation of $\textsc{roll}(\hat{K}, l)^\top \hat{Q}$ takes $O(d_k^2 T)$ time. With our choice of $k = O(\log(T))$, Equation \ref{letstopk3} takes  $O(d_k^2 T \log(T))$ time. Nevertheless, since Equation \ref{letstopk2} requires iterating over all $T-1$ lags $l \in [1, T-1]$, each of which costs $O(d_k^2 T)$, the total complexity is $O(d_k^2 T^2)$. We hereby present how to alleviate the computation  in Equation \ref{letstopk2} via FFT, thereby resulting in the accelerated complexity of $O(d_k^2 T \log(T))$.
This is enabled via the Cross-correlation Theorem \citep{cross_corr_theorem}, which, given two finite discrete time series $\{ \Xx_t\}$ and $\{ \Yy_t\}$, permits the sliding inner product $(\Xx \star \Yy)(l) = \sum_{t=1}^{T} \Xx_{t-l} \Yy_t  $ of different lag values $l\in[0,T-1]$ being computed efficiently via FFT as:
\begin{align}
\label{fft_trick}
\begin{split}
    \Ss_{\Xx \Yy}(f) &= \Ff(\Xx_t) \Ff^*(\Yy_t) = \int_{-\infty}^{+\infty} \Xx_t e^{-i2 \pi t f } dt \overline{\int_{-\infty}^{+\infty} \Yy_t e^{-i2 \pi t f } dt} \\
    (\Xx \star \Yy)(l) &= \Ff^{-1}( \Ss_{\Xx \Yy}(f) ) =  \int_{-\infty}^{+\infty} \Ss_{\Xx \Yy}(f) e^{i 2 \pi f l} df,
\end{split}
\end{align}
for $l \in [0, T-1]$, where $\Ff$ and $\Ff^{-1}$ are FFT and FFT inverse, and $*$ is the conjugate operation.
 Particularly, given $\bar{K}, \bar{Q} \in \Br^{T \times d_k}$, we first compute $\Ff(\bar{K}), \Ff(\bar{Q}) \in \Br^{(T/2+1 )\times d_k}$ in the frequency domain. Let $\Ff(.)_i$ be the $i^{th}$ column of these  FFTs. We then compute $\Ff(\bar{K})_i \Ff^*(\bar{Q})_j $ for all $i, j \in [1, d_k]$. Finally, the inverse FFTs of these products would give $\Ff^{-1}\big( \Ff( \bar{K})_i  \Ff^*( \bar{Q})_j \big) = \big[ \big( \textsc{roll}( \bar{K}, 0)^\top \bar{Q}\big)_{ij},\big( \textsc{roll}( \bar{K}, 1)^\top \bar{Q}\big)_{ij}, ..., \big( \textsc{roll}( \bar{K}, T-1)^\top \bar{Q}\big)_{ij}  \big]$ for $i, j \in [1, d_k]$. Thus, we can gather data to obtain $\textsc{roll}( \bar{K}, l)^T \bar{Q}$ for all $l\in[0, T-1]$. As each of  FFT and inverse FFT  takes $O(T \log(T)$,  CAB  achieves the $O(d_k^2 T \log(T))$ complexity. We note that the cross-correlation computation required by CAB is more complicated and strictly subsumes auto-correlation and the invoked Cross-correlation Theorem  is more generalized version of the Wiener–Khinchin Theorem, as used by \citep{wu2022autoformer}  for auto-correlation computation.

\textbf{Differences Compared to Autoformer.} Since the CAB aims to capture the lagged cross-correlation, which is  relevant to yet more generalized than the auto-correlation  in Autoformer, we believe it is crucial to emphasize the main differences. First, Autoformer overall is a decomposed encoder-decoder architecture  proposed for long-term forecasting, so its auto-correlation module is specifically designed to work with series seasonalities extracted from various series decomposition steps of Autoformer. On the other hand, CAB ensures flexibility with any input series representation 
 by deploying normalization step and  learnable temperature coefficient $\lambda$ reweighting the correlation matrices. Second, while Autoformer computes purely auto-correlation scores and aggregates their exact values  for TopK ,  CAB computes cross-correlation matrices and aggregates the absolute values of such entries for TopK  in Equation \ref{letstopk2} (as correlation can stem from either positive or negative correlation). Finally, to facilitate robustness to different input series representation, CAB adopts learnable weights $\lambda$ in TopK operation, which balances between auto-correlation and cross-correlation, and $\beta$  in  sub-series aggregation, which balances between instantaneous and lagged cross-correlation.

\subsubsection{Mixture-of-head Attention}

For seamless integration of CAB with a broad class of encoder-only Transformer architectures using  multi-head attention component (e.g. \citep{NIPS2017_3f5ee243, liu2022nonstationary, Du_2023, zhou2021informer}), we propose mixture-of-head attention  that leverages a mixture of both temporal attentions and correlated attentions. mixture-of-head attention  modifies multi-head attention  (Equation \ref{multihead_attention}) to also incorporate CAB as follows:
\begin{align}
\label{mixture_attention}
\begin{split}
    &\textsc{Mixture-of-head-Attention}(X) = \textsc{Concat}(head_1, head_2, ..., head_h) W^O \\
    \text{wh}&\text{ere } head_i = \begin{cases} \textsc{Temporal-Attention}(X W^Q_i, X W^K_i, X W^V_i), \text{ if } i \leq m\\
        \textsc{Correlated-Attention}(X W^Q_i, X W^K_i, X W^V_i), \text{ otherwise } 
    \end{cases},
\end{split}
\end{align}
where $m$ is a threshold hyperparameter that controls the split between temporal attention and correlated attention. This uncomplicated modification to the base architecture of multi-head attention allows CAB to be flexibly plugged into a wide range of existing and future  Transformers. 

%% file: sections/experiments.tex
\section{Experiments}
\label{sec_exp}

As CAB is a plug-in attention for encoder-only Transformer architectures, we extensively experiment on three mainstream MTS non-predictive tasks including imputation, anomaly detection and classification on real-world datasets. 
Ablation studies are provided in Appendix \ref{appen_ablation} analyzing the efficiency of each component of CAB. 
While  focusing on non-predictive tasks, we provide  preliminary results on MTS long-term forecasting  in Appendix \ref{appen_forecast}.  
Run-time analysis is presented  in Appendix \ref{appen_runtime}.
\begin{table}[H]
\caption{Dataset Summary}
\label{dataset_summary}
\resizebox{\linewidth}{!}{%
\begin{tabular}{S|S|S|S} \toprule
\small
    {MTS Analysis Tasks} & {Benchmarking Datasets} &{Metrics}& {Sequence Length} \\ 
    \midrule
    \text{Imputation}  & {ETTm1, ETTm2, ETTh1, ETTh2, Electricity, Weather} & {MSE, MAE} & 96 \\
     \midrule
    \text{Anomaly Detection}  & {SMD, MSL, SMAP, SWaT, PSM} & {Precision, Recall, F1-score (\%)} & 100 \\
    \midrule
    \text{Classification}  & {UEA (10 subsets)} & {Accuracy (\%)} & {29-1751} \\
    \bottomrule
\end{tabular}
}
\end{table}
\textbf{Experiment Benchmarks.}  Following \citep{zhou2021informer, wu2023timesnet, classi_benchmark1}, we extensively benchmark  over the following real-world datasets: ETTh1 and ETTh2 (Electricity Transformer Temperature-hourly) \citep{zhou2021informer}, ETTm1 and ETTm2 (Electricity Transformer Temperature-minutely) \citep{zhou2021informer}, Electricity \citep{misc_electricityloaddiagrams20112014_321}, Weather \citep{weather_data}, SMD \citep{SMD1}, MSL \citep{MSL1}, SMAP \citep{MSL1}, SWaT \citep{SWAT1}, PSM \citep{PSM1} and UEA Time Series Classification Archive \citep{bagnall2018uea}. A summary of the datasets for benchmark is given in  Table \ref{dataset_summary}.

\textbf{Baselines.} We compare with \emph{TimesNet} \citep{wu2023timesnet}\footnote{ While  results of TimesNet on forecasting and imputation are reproducible, we  cannot recover its state-of-the-art results, from  their released code, on anomaly detection and classification. We report here the results on such two tasks obtained from their released implementation and note that  the relative  ranking of baselines remains the same as in TimesNet benchmark \citep{wu2023timesnet}, i.e. TimesNet is the best among the previous baselines.}, the current state-of-the-art deep learning model on these three tasks (though not being Transformer-based), \emph{DLinear} \citep{zeng2022transformers}, and the prevalent Transformer-based models including  vanilla \emph{ Transformer} \citep{NIPS2017_3f5ee243}, \emph{Nonstationary Transformer} \citep{liu2022nonstationary}, which has been  shown to consistently achieve competitive results on a variety of tasks,   \emph{FEDformer} \citep{fedformer1}, and \emph{Autoformer} \citep{wu2022autoformer}. In fact, Nonstationary Transformer and FEDformer are    the state-of-the-art Transformer-models for respectively imputation and anomaly detection in the recent benchmarks \citep{wu2023timesnet}. For classification, we  also consider \emph{Flowformer} \citep{Wu2022FlowformerLT}, the state-of-the-art Transformer-model.

\textbf{Our Models.} 
We integrate our CAB (through the mixture-of-head attention framework) into the two representative base models: \emph{Transformer} \citep{NIPS2017_3f5ee243} and  \emph{Nonstationary Transformer} \citep{liu2022nonstationary}, whose attention mechanisms are reviewed in Section \ref{sec_background}. Implementation details and hyper-parameter settings are given in Appendix \ref{appen_implementation}.



\subsection{Imputation}
\label{sec_impute}

\textbf{Setup.} Due to uncertainties of natural processes and malfunction of sensors, partially missing data is common in MTS data, thereby hindering direct adoption of off-the-shelf models as well as posing challenges to downstream task analysis. Consequently, MTS imputation is crucial and has gathered much research interest \citep{LOPEZ2021104794}. To exemplify real-world scenario commonly facing data missing problem, we consider the following six datasets from electricity and weather domain for benchmark: ETTh1 and ETTh2 (Electricity Transformer Temperature-hourly) \citep{zhou2021informer}, ETTm1 and ETTm2 (Electricity Transformer Temperature-minutely) \citep{zhou2021informer}, Electricity \citep{misc_electricityloaddiagrams20112014_321} and Weather \citep{weather_data}. Each original dataset is firstly split into three sets of training set, validation set, and test set respectively with ratio $60\%, 20\%$ and $20\%$.  Time-series data is generated by selecting every 96 consecutive steps as a sample.
To test the models under different missing data rate, we randomly mask the time points with the ratio of $\{12.5\%, 25\%, 37.5\%, 50\%\}$. 
 We adopt the mean square error (MSE) and mean absolute error (MAE) as the metrics.

\textbf{Results.} The results are depicted in Table \ref{imputation_table}. 
The base Transformers when combined with CAB achieve competitive results. Particularly, 
Nonstationary+CAB and Transformer+CAB   improve over Nonstationary and Transformer in respectively five and four datasets out of the total of six datasets. Nonstationary+CAB achieves state-of-the-art results surpassing TimesNet on  five datasets.



\begin{table}[H]
\caption{Imputation task over six datasets. The missing data rate is $\{12.5\%, 25\%, 37.5\%, 50\%\}$ and series length is 96. We highlight \textcolor{red}{the best} results and \textcolor{blue}{the second best} results.}
\label{imputation_table}
\resizebox{\linewidth}{!}{%
\begin{tabular}{S|S|SS | SS |SS |SS |SS | SS | SS | SS} \toprule
    {Datasets} & {Mask Ratio} &\multicolumn{2}{c}{\makecell{TimesNet \\ \citep{wu2023timesnet}}} & \multicolumn{2}{c}{\makecell{Nonstationary \\ \citep{liu2022nonstationary}}} & \multicolumn{2}{c}{ \makecell{Nonstationary+CAB \\ \textbf{(Ours)} }} & \multicolumn{2}{c}{ \makecell{Transformer \\ \citep{NIPS2017_3f5ee243}} } & \multicolumn{2}{c}{\makecell{Transformer+CAB\\\textbf{(Ours)}}}  & \multicolumn{2}{c}{ \makecell{FEDformer \\ \citep{fedformer1}} } & \multicolumn{2}{c}{ \makecell{DLinear \\ \citep{zeng2022transformers}} } & \multicolumn{2}{c}{ \makecell{Autoformer \\ \citep{wu2022autoformer}} } \\ 
    \midrule
    {} & {} & {MSE}&{MAE}& {MSE}&{ MAE}& {MSE}&{ MAE}& {MSE}&{ MAE} & {MSE}&{ MAE} & {MSE}&{ MAE}& {MSE}&{ MAE} & {MSE}&{ MAE} \\
    \midrule
    \text{ETTm1}  & 12.5\%  &  \textcolor{blue}{0.019} & \textcolor{blue}{0.092} &   {0.026} & {0.107} &       \textcolor{red}{0.018} & \textcolor{red}{0.087}  &  0.023 & 0.105 & 0.022 & 0.104 & 0.035 & 0.135  & 0.058 & 0.162  & {0.034} & {0.124}\\
    \text{ETTm1}  & 25\%   &  \textcolor{blue}{0.023} & \textcolor{blue}{0.101}  &   {0.032} & {0.119}  &    \textcolor{red}{0.02} & \textcolor{red}{0.097}  &  0.030  & 0.121 & 0.031 & 0.123 & 0.052 & 0.166 & 0.080 & 0.193  & {0.046} & {0.144}\\
    \text{ETTm1}  & 37.5\%  &   \textcolor{red}{0.029} & \textcolor{red}{0.111}  &   {0.039} & {0.131} &    \textcolor{blue}{0.030} & \textcolor{blue}{0.112} &  0.037 & 0.135 &  0.039 & 0.140 & 0.069 & 0.191 &  0.103 & 0.219 & {0.057} & {0.161}\\
    \text{ETTm1} & 50\%   &  \textcolor{red}{0.036} & \textcolor{red}{0.124} &   {0.047} & {0.145}    &     \textcolor{blue}{0.037} & \textcolor{blue}{0.125} &  0.045 & 0.148 & 0.050 & 0.157 & 0.089 & 0.218 & 0.132 & 0.248 &  {0.067} & {0.174}\\
    \textbf{Average} &   &   \textcolor{blue}{0.027} & \textcolor{blue}{0.106}  &   {0.036} & {0.126} &\textcolor{red}{ 0.026}    & \textcolor{red}{0.105}   &  0.034 & 0.127 & 0.036 & 0.131 &0.062 & 0.177 &0.093 &0.206 & {0.051} & {0.150}\\
    \midrule
    \text{ETTm2}  & 12.5\%  &  \textcolor{blue}{0.018} & \textcolor{blue}{0.080}  &   {0.021} & {0.088} &    \textcolor{red}{0.016} & \textcolor{red}{0.076} &  0.125 & 0.264 & 0.136 & 0.271 & 0.056 & 0.159  & 0.062 & 0.166 & {0.023} & {0.092}\\
    \text{ETTm2}  & 25\%  &   \textcolor{blue}{0.020} & \textcolor{blue}{0.085}  &  {0.024} & {0.096}  &    \textcolor{red}{0.018} & \textcolor{red}{0.082} & 0.195  & 0.323 & 0.152 & 0.288 & 0.080 & 0.195  &  0.085 & 0.196 & {0.026} & {0.10}\\
    \text{ETTm2}  &37.5\%  &   \textcolor{red}{0.023} & \textcolor{red}{0.091} &  {0.027} & {0.103}   &  \textcolor{blue}{0.024} & \textcolor{blue}{0.092} &  0.217 &  0.343 & 0.179 & 0.312  &  0.110 & 0.231&0.106 & 0.222 & {0.030} & {0.108}\\
    \text{ETTm2}  & 50\%  &  \textcolor{red}{0.026} & \textcolor{red}{0.098} &   {0.030} & {0.108}  &    \textcolor{blue}{0.027} & \textcolor{blue}{0.099}  &  0.257 & 0.378 & 0.211 & 0.340 &  0.156 & 0.276 &  0.131 & 0.247 &  {0.035} & {0.119}\\
    \textbf{Average} &   &   \textcolor{blue}{0.022} & \textcolor{blue}{0.088}  &  {0.026} & {0.099}   & \textcolor{red}{0.021} & \textcolor{red}{0.087} &  0.199 & 0.327 &  0.170 & 0.303 & 0.101 & 0.215 & 0.096 & 0.208 & {0.029} &  {0.105}\\
    \midrule
    \text{ETTh1} & 12.5\%  &   \textcolor{blue}{0.057} & \textcolor{blue}{0.159}  &  {0.060} & {0.165}   &  \textcolor{red}{0.047} & \textcolor{red}{0.148} &  0.063 & 0.178 & 0.070 & 0.189 & 0.070 & 0.190 &  0.151 & 0.267 & {0.074} & {0.182}\\
    \text{ETTh1} & 25\%   &   \textcolor{blue}{0.069} & \textcolor{blue}{0.178}   &     {0.080} & {0.189}   &   \textcolor{red}{0.064} & \textcolor{red}{0.171} &  0.089 & 0.212 & 0.098 &0.223 & 0.106  & 0.236 & 0.180 & 0.292  &{0.090} & {0.203}\\
    \text{ETTh1} & 37.5\%   &   \textcolor{red}{0.084} & \textcolor{red}{0.196}   &  { 0.102} & {0.212}   &   \textcolor{blue}{0.085} & \textcolor{blue}{0.195} &  0.115 & 0.242 & 0.137 & 0.264  &  0.124 & 0.258 & 0.215 & 0.318 & {0.109} & {0.222}\\
    \text{ETTh1} & 50\%    &   \textcolor{red}{0.102} & \textcolor{red}{0.215}  &   { 0.133} & {0.240}   &   \textcolor{blue}{0.106} & \textcolor{blue}{0.216} &  0.140 & 0.270 & 0.162 & 0.286 & 
 0.165 & 0.299 &  0.257 & 0.347 & {0.137} & {0.248}\\
    \textbf{Average} &   &      \textcolor{blue}{0.078 } & \textcolor{blue}{0.187}   &   {0.094} & {0.201}   & \textcolor{red}{0.076} & \textcolor{red}{0.182} & 0.102 & 0.226 &  0.117 & 0.241 & 0.117 & 0.246 &   0.201 & 0.306 & {0.103} & {0.214}\\
    \midrule
    \text{ETTh2} & 12.5\%   &   \textcolor{blue}{0.040} & \textcolor{blue}{0.130}   &   {0.042} & {0.133}    &  \textcolor{red}{0.039} & \textcolor{red}{0.129} &  0.205  & 0.329 &  0.212 & 0.354 &  0.095 & 0.212& 0.100 & 0.216 & {0.044} & {0.138}\\
    \text{ETTh2} & 25\%   &   \textcolor{blue}{0.046} & \textcolor{blue}{0.141}   &    {0.049} & {0.147}   & \textcolor{red}{0.044} & \textcolor{red}{0.139} &  0.283  & 0.397 & 0.228 & 0.355  &  0.137  & 0.258  &  0.127 & 0.247 &  {0.050} & {0.149}\\
    \text{ETTh2} & 37.5\%   &   \textcolor{blue}{0.052} & \textcolor{blue}{0.151}   &   {0.056} & {0.158}   & \textcolor{red}{0.051} & \textcolor{red}{0.150} &  0.285 & 0.392 & 0.265 & 0.378 & 0.187 &  0.304 & 0.158 & 0.276 & {0.060} &{0.163}\\
    \text{ETTh2} & 50\%  & \textcolor{blue}{0.060}& \textcolor{blue}{0.162}   &  {0.065} & {0.170}   &  \textcolor{red}{0.059} & \textcolor{red}{0.160} &  0.327 & 0.418 & 0.319  & 0.415 &  0.232 & 0.341 & 0.183 & 0.299 & {0.068} & {0.173}\\
    \textbf{Average} &   &   \textcolor{blue}{0.049} & \textcolor{blue}{0.146}   &    {0.053} & {0.152}   &  \textcolor{red}{0.048} & \textcolor{red}{0.145} & 0.275 & 0.384 &  0.256 & 0.376   & 0.163 & 0.279 &  0.142 & 0.259 & {0.055} & {0.156}\\
    \midrule
    \text{Electricity} & 12.5\%  &  \textcolor{blue}{0.085} & \textcolor{blue}{0.202}  &   {0.093} & {0.210}   &  \textcolor{red}{ 0.081} & \textcolor{red}{0.198}  &  0.148  & 0.276 & 0.143  & 0.269  & 0.107 & 0.237 &  0.092 & 0.214 & {0.089} & {0.210}\\
    \text{Electricity} & 25\%   &   \textcolor{blue}{0.089} & \textcolor{blue}{0.206}   &   {0.097} & {0.214}    &    \textcolor{red}{0.087} & \textcolor{red}{0.204} &  0.161  & 0.285 &  0.165  & 0.283 &  0.120 & 0.251 & 0.118 & 0.247  &  {0.096} & {0.220}\\
    \text{Electricity} & 37.5\%   &   \textcolor{blue}{0.094} & \textcolor{blue}{0.213}   &    {0.102} & {0.220}   &  \textcolor{red}{0.093} & \textcolor{red}{0.209} &  0.170 & 0.292 &  0.168 & 0.290  &  0.136 & 0.266 & 0.144 & 0.276 & {0.104} & {0.229}\\
    \text{Electricity} & 50\%  &   \textcolor{blue}{0.100} & \textcolor{blue}{0.221}   &   {0.108} & {0.228}   &  \textcolor{red}{0.098} & \textcolor{red}{0.215} &  0.177 & 0.296 & 0.173 & 0.295 &   0.158 & 0.284  &  0.175 & 0.305 & {0.113} & {0.239}\\
    \textbf{Average} &   &   \textcolor{blue}{0.092} & \textcolor{blue}{0.210}  &  { 0.100} & {0.218}    & \textcolor{red}{0.089}   & \textcolor{red}{0.207} &  0.164 & 0.287 &  0.162 & 0.284  & 0.130 & 0.259  & 0.132 & 0.260 & {0.101} & {0.225}\\
    \midrule
    \text{Weather} & 12.5\%  &   \textcolor{red}{0.025} & \textcolor{red}{0.045}  &   {0.027} & {0.051}   &  \textcolor{blue}{ 0.026} & {0.050} &  0.034  & 0.090 & 0.033 & 0.082  & 0.041 & 0.107 &  0.039 & 0.084  & {0.026} & \textcolor{blue}{0.047}\\
    \text{Weather} & 25\%  &   \textcolor{blue}{0.029} & \textcolor{red}{0.052}    &   {0.029} & {0.056} &    \textcolor{red}{0.029} & {0.056} &  0.036  &  0.089 & 0.034 & 0.085 &  0.064 & 0.163  & 0.048 & 0.103 & {0.030} & \textcolor{blue}{0.054}\\
    \text{Weather} & 37.5\%  &    \textcolor{red}{0.031}  & \textcolor{red}{0.057} &   {0.033} & {0.062}  &    {0.034} & {0.064} &  0.038  & 0.091 & 0.038 & 0.089  &  0.107 & 0.229  &  0.057 & 0.117 & \textcolor{blue}{0.032} & \textcolor{blue}{0.060}\\
    \text{Weather} & 50\%   &   \textcolor{red}{0.034} & \textcolor{red}{0.062} &   \textcolor{blue}{0.037} & {0.068}  &    {0.041} & {0.074} &   0.042 & 0.095 & 0.046 & 0.105  & 0.183 & 0.312 & 0.066 & 0.134 &  \textcolor{blue}{0.037} &\textcolor{blue}{0.067}\\
    \textbf{Average} &   &   \textcolor{red}{0.030} & \textcolor{red}{0.054}   &    {0.032} & {0.059}   &  {0.032} & {0.061} &  0.038 &  0.091 & 0.038 & 0.090  &  0.099 & 0.203 & 0.052 & 0.110 & \textcolor{blue}{0.031} & \textcolor{blue}{0.057}\\
    \bottomrule
\end{tabular}
}
\end{table}

\subsection{Anomaly Detection}

\textbf{Setup.} Anomalies are inherent in large-scale data and can be caused by noisy measurements. We consider the five datasets vastly used for anomaly-detection benchmarks: SMD \citep{SMD1}, MSL \citep{MSL1}, SMAP \citep{MSL1}, SWaT \citep{SWAT1} and PSM \citep{PSM1}. We then follow \citep{xu2022anomaly, NEURIPS2020_97e401a0} for pre-processing data that generates a set of sub-series via non-overlapped sliding window, and set the series length to 100. The original datasets SMD, MSL, SMAP, SWaT and PSM are splitted into collections of training set, validation set and test set following  \citep[Appendix K]{xu2022anomaly}. We adopt Precision, Recall and F1-score (all in \%) as the metrics, where higher values correspond to better performance. 

\textbf{Results.} From Table \ref{anomaly_table}, our model Nonstationary+CAB achieves the best average F1-score, surpassing TimesNet. Furthermore, CAB consistently and significantly improves the precision and F1-score, which is the more favorable metrics for balancing precision and recall, of the base Transformers.



\begin{table}[H]
\caption{Anomaly detection task over five datasets. We report the Precision (P), Recall (R) and F1-score (F1)- the harmonic mean of precision and recall, and highlight \textcolor{red}{the best} results and \textcolor{blue}{the second best} results. }
\label{anomaly_table}
\resizebox{\linewidth}{!}{%
\begin{tabular}{S|S S S|S S S|S S S|S S S|S S S|S S S|S S S|S S S} \toprule
    {Datasets}  & \multicolumn{3}{c}{\makecell{TimesNet \\ \citep{wu2023timesnet}}} & \multicolumn{3}{c}{\makecell{Transformer \\ \citep{NIPS2017_3f5ee243}} }&  \multicolumn{3}{c}{\makecell{Transformer+CAB\\\textbf{(Ours)}}} & \multicolumn{3}{c}{\makecell{Nonstationary \\ \citep{liu2022nonstationary}}} & \multicolumn{3}{c}{ \makecell{Nonstationary+CAB \\ \textbf{(Ours)} }} & \multicolumn{3}{c}{\makecell{FEDformer \\ \citep{fedformer1}}} & \multicolumn{3}{c}{\makecell{DLinear \\ \citep{zeng2022transformers}} } &  \multicolumn{3}{c}{\makecell{Autoformer \\ \citep{wu2022autoformer}} }  \\ 
    \midrule
    {} & {P} & {R} & {F1} & {P} & {R} & {F1}& {P} & {R} & {F1}& {P} & {R} & {F1}& {P} & {R} & {F1}& {P} & {R} & {F1}& {P} & {R} & {F1} & {P} & {R} & {F1}\\
    \midrule
    \text{SMD}  & {87.88} & {81.54} & {84.59} & {83.58} & { 76.13 }& {79.56}& {78.36} & {65.25}& {71.20} & \textcolor{blue}{88.33} & {81.21} & {84.62}& \textcolor{red}{90.43} & {82.33} & \textcolor{red}{86.19 } & {87.95 } & \textcolor{red}{82.39 } & {85.08} & {83.62} & { 71.52} & {77.10}  & {88.06} & \textcolor{blue}{82.35} & {85.11}\\
    \text{MSL}  &  \textcolor{blue}{89.55} & {75.29} & \textcolor{blue}{81.80}& { 71.57} & \textcolor{blue}{87.37 } & {78.68} & \textcolor{red}{89.70} & {73.66} &  {80.90}& {68.55} & \textcolor{red}{  89.14} & {77.50} & {88.02} & {72.83}& {79.71 } & {77.14} & {80.07} & {78.57} & {84.34} & {85.42 } & \textcolor{red}{ 84.88} & {77.27}& {80.92} & {79.05}\\ 
    \text{SMAP} & {90.05} & {56.54} & {69.46}& { 89.37} & {57.12} & {69.70} & \textcolor{blue}{90.86} & \textcolor{red}{61.87}&  \textcolor{red}{73.79}& {89.37} & \textcolor{blue}{ 59.02} & \textcolor{blue}{71.09 }&  {90.27} & {57.3} & {70.10 } & { 90.47} & { 58.10} & {70.76} & \textcolor{red}{92.32 } & {55.41 } & { 69.26 } & {90.40}& {58.62}& {71.12}\\
    \text{SWaT} & \textcolor{blue}{90.95}& {95.42} & \textcolor{blue}{93.13}& {68.84} & \textcolor{blue}{96.53} &  {80.37} & \textcolor{red}{99.67} & {68.89}  & {81.47} & {68.03 } & \textcolor{red}{ 96.75} & {79.88}& {90.55} & {95.41} & {92.92} & { 90.17} & { 96.42}& \textcolor{red}{93.19} & { 80.91} & {95.30} & {87.52} & {89.85} &  {95.81} & {92.74} \\
    \text{PSM}  & \textcolor{blue}{98.51} & {96.29} & \textcolor{blue}{97.39}& {62.75} & {96.56 } & {76.07} & \textcolor{red}{99.34} & {82.92} & {90.39}& { 97.82} & {96.76} & {97.29} & {98.25} & {96.13}& \textcolor{red}{97.58} & { 97.31} & \textcolor{red}{97.16 }  & {97.23} & {98.28 } & {89.26} & {93.55 } & \textcolor{blue}{99.08} & {88.15} & {93.29} \\
    \midrule
    \textbf{Average} &  {91.39} & {81.02} & \textcolor{blue}{85.27} & {75.22} & {82.74}& {76.88}  & \textcolor{red}{91.59} & {70.52} & {79.55} & {82.42} & \textcolor{red}{84.06} & {82.08}  & \textcolor{blue}{91.50} & {80.80}& \textcolor{red}{85.30} & {88.61 } & \textcolor{blue}{82.83}& {84.97} & {87.89} & {79.38}  & {82.46} & {88.93} & {81.17} & {84.262}\\
    \bottomrule
\end{tabular}
}
\end{table}

\subsection{Classification}

\textbf{Setup.} We select ten datasets from the UEA Time Series Classification Archive \citep{bagnall2018uea} following \citep{wu2023timesnet}. These cover health care, audio recognition, transportation and other practical applications. The datasets are pre-processed similarly to \citep[Appendix A]{classi_benchmark1} that assigns different series length for different subsets. We adopt the accuracy (\%) as the metrics.

\textbf{Results.} As shown in Table \ref{classification_table}, our model Transformer+CAB achieves the best overall result surpassing  TimesNet. Moreover, CAB demonstrates consistent performance improvement when combined with either Transformer or Nonstationary Transformer.

\begin{table}[H]
\caption{Classification task task over 10 datasets from UEA. The accuracies (\%) are reported. We highlight \textcolor{red}{the best} results and \textcolor{blue}{the second best} results.}
\label{classification_table}
\resizebox{\linewidth}{!}{%
\begin{tabular}{S|S|S|S|S|S|S|S|S|S} \toprule
    {Datasets}  &{\makecell{TimesNet \\ \citep{wu2023timesnet}}} & {\makecell{Transformer \\ \citep{NIPS2017_3f5ee243}} }&  {\makecell{Transformer+CAB\\\textbf{(Ours)}}} &{\makecell{Nonstationary \\ \citep{liu2022nonstationary}}} & { \makecell{Nonstationary+CAB \\ \textbf{(Ours)} }} &{\makecell{FEDformer \\ \citep{fedformer1}}} & {\makecell{DLinear \\ \citep{zeng2022transformers}} } & {\makecell{Flowformer \\ \citep{Wu2022FlowformerLT}} } &  {\makecell{Autoformer \\ \citep{wu2022autoformer}} }  \\ 
    \midrule
    \text{Ethanol }  &  {28.14} & {26.24}& \textcolor{blue}{31.94} & {25.10}  & {25.10} &  {28.90} & {27.00} & \textcolor{red}{33.08} & { 27.38 }\\
    \text{FaceDetection}  &  {67.31}  &  {67.93}  & \textcolor{red}{ 71.11} & {68.70}  &   \textcolor{blue}{69.40} & {68.55} & { 67.25} &{67.08}& { 54.63 } \\ 
    \text{Handwriting}  &  {29.88} & {29.53}   & {29.06} & \textcolor{red}{31.41}  & \textcolor{blue}{30.12} & {18.47}&  {18.94 } & {27.18} & { 13.18 } \\
    \text{Heartbeat}  &  {74.15} & \textcolor{blue}{75.12}   &  \textcolor{red}{75.12} & {72.20}  & {72.20} & {75.12} & { 70.73} & {72.68} & { 69.76} \\
    \text{JapaneseVowels}   &  {97.57} &  {97.03}    & \textcolor{blue}{97.84} & {96.22}  &  {95.68} & {96.76}& {94.86 } &\textcolor{red}{98.65} & { 94.86}  \\
    \text{PEMS-SF}  & \textcolor{red}{89.02} & {78.03}  & \textcolor{blue}{86.71}& {82.66}  &{75.14}& {86.71} & { 80.35 } &{ 86.71} & { 82.66 }  \\
    \text{SCP1}   & {91.13} & \textcolor{blue}{91.13} & \textcolor{red}{91.47} & {83.28}  &  {82.94} & { 57.00}&{ 88.05} & {89.08}  & {  59.39}  \\
    \text{SCP2}   & {52.78} & {53.89 }  &  \textcolor{red}{56.11}&  {50.00}  &  \textcolor{blue}{55.55}& { 49.44}&{52.78 } & {54.44}  & { 53.89 }  \\
    \text{SpokenArabic} &  {98.68} & {98.45}    & \textcolor{red}{99.05}  & {98.82 }  &  {98.91} &  { 98.32}&{ 96.54}& \textcolor{blue}{98.95}  & { 98.82 }  \\
    \text{UWaveGesture}    & \textcolor{red}{86.88} & {86.25}  &  {85.94} & {81.56}  & {85.94} &  { 44.06}&{ 81.25} & \textcolor{blue}{86.88}  & {45.63  }  \\
    \midrule
    \textbf{Average}  & \textcolor{blue}{71.49}  & {70.36} & \textcolor{red}{72.44} & {69.00}  & {69.10}  &  { 62.33}&{ 67.78} & {71.47}  & {  60.02}\\
    \bottomrule
\end{tabular}
}
\end{table}

%% file: sections/conclusion.tex
\section{Conclusion and Future Work}

In this paper, we proposed the novel correlated attention block (CAB) that can efficiently learn the cross-correlation between variates of MTS data, and  be seamlessly plugged into existing Transformer-based models for performance improvement. The modularity of CAB, which could be flexibly plugged into  follow-up Transformer-architectures for efficiency gain, and the methodology behind our design of CAB, which is the first attention mechanism that aims to capture lagged cross-correlation in the literature,  will greatly benefit future work on time series Transformers. Extensive experiments on imputation, anomaly detection and classification demonstrate the benefits of CAB for improving base Transformers, and result in state-of-the-art models for respective tasks. For future work, we will extend the design of CAB to be integrated into encoder-decoder Transformer-architectures for improving performance in MTS predictive tasks.

%% file: sections/appendix.tex


\section{Implementation Details}
\label{appen_implementation}

The models are  implemented in PyTorch and experiments are run on a NVIDIA-SMI v100 GPU with 100Gb of storage. The learnable parameters are initialized as $\tau = 1$, and $\lambda = \beta = 1/2$. For MTS datasets with dimension $d < 70$, we set $d_{model} = d_k = 64$; otherwise, we use $d_{model}= d_k = 128$. For PEMS-SF, we set $d_{model} = d_k = 180$. If $d_k < 100$, we fix $\lambda = 1/2$ in Equation \ref{letstopk2}. The hyperparameter $c$ in the $k = c \lceil \log(T) \rceil$ of TopK is chose among $c \in \{1, 2, 3\}$, where we pick out the value with best result.
For the mixture-of-head attention, we use $h=16$ and $m=8$, i.e. 8 heads for temporal attentions and 8 heads for correlated attentions. The loss for classification is entropy, while the loss for imputation and anomaly detection is MSE. 
We use ADAM for training with the default  hyperparameter configuration. Batch size is set to 16 for imputation and classification, and 128 for anomaly detection. The number of epochs is set to 30.  If the validation loss does not decreases for 10 epochs, the training is stopped.

\section{Ablation Studies}
\label{appen_ablation}

The main components of CAB are the lagged cross-correlation filtering step (Equation \ref{letstopk2}), which involves the learnable $\lambda$ to untangle auto-correlation and cross-feature correlation, and score aggregation step (Equation \ref{letstopk3}), which involves the learnable $\beta$ to balance instantaneous cross-correlation and lagged cross-correlation. A key component for the seamless integration of CAB into base Transformers is the mixture-of-head attention (MOHA) that utilizes a mixture of temporal and correlated attention heads. To test each of the aforementioned components, we use the same experimental setting for classification as in Section \ref{sec_exp}, and consider the following ablation versions of Transformer+CAB, the best performing model for this task:

\begin{itemize}
    \item \textbf{pure-CAB-Transformer}: \emph{(testing MOHA)} In this model, we simply replace the self-attention of vanilla Transformer with the most basic correlated attention setting where lagged cross-correlation filtering step is disabled, i.e. no learning of $\lambda$, and $\beta$ is set to 0; hence score aggregation steps just returns the instantaneous cross-correlation. In short, self-attention is now replaced by:
    \begin{align*}
        \textsc{Correlated-Attention}(Q, K, V) = V \textsc{softmax}(\frac{1}{\tau} \hat{K}^\top \hat{Q}),
    \end{align*}
    to test how this simple mechanism can take place of self-attention.
    \item \textbf{static-Transformer+CAB}: \emph{(testing lagged cross-correlation filtering)} In this model, we use 8 self-attention heads and 8 correlated attention heads, and enable back the lagged cross-correlation filtering, yet hard-fix $\lambda = \beta = 1/2$, i.e. there is no learning of these parameters. This is to test the simplified lagged cross-correlation filtering's efficiency. 
    \item \textbf{$\lambda$-Transformer+CAB}: \emph{(testing $\lambda$)} This is the same as static-Transformer+CAB except that we now allow $\lambda$ to be learnable parameter. $\beta$ is fixed to $1/2$.
    \item \textbf{$\beta$-Transformer+CAB}:  \emph{(testing $\beta$)} This is the same as static-Transformer+CAB except that we now allow $\beta$ to be learnable parameter.  $\lambda$ is fixed to $1/2$.
    
\end{itemize}

 \begin{table}[H]
\caption{Ablation studies where the original  model Transformer+CAB is compared to the variants with  different disabled components.}
\label{ablation_table}
\resizebox{\linewidth}{!}{%
\begin{tabular}{S|S|S|S|S|S} \toprule
\small
    {Dataset/Method}  &{Transformer+CAB}&{pure-CAB-Transformer} &{static-Transformer+CAB} & { $\lambda$-Transformer+CAB}& { $\beta$-Transformer+CAB}\\ 
    \midrule
    \text{Ethanol}  & 31.94& 30.79 & 28.14 & 29.28 &  30.42\\
    \text{FaceDetection}   &  71.11  & 68.81 & 69.67 & 70.20 &   70.83 \\ 
    \text{Handwriting}     & 29.06 &  20.94 & 23.88 & 24.35 & 29.06 \\
    \text{Heartbeat}  &  75.12  & 72.68 & 75.61 & 74.15 & 75.12\\
    \text{JapaneseVowels}   &  97.84 &  96.76 & 95.68 & 95.68 &  97.84\\
    \text{PEMS-SF}  & 86.71& 72.83  & 79.77 & 76.88&  84.39\\
    \text{SCP1}  & 91.47 &  86.01 & 90.44 & 88.74 & 91.13\\
    \text{SCP2}  & 56.11  &  58.89 & 57.22 & 55.00 &  55.00\\
    \text{SpokenArabic}  & 99.05  &  98.45 & 98.54 & 99.27 & 99.05 \\
    \text{UWaveGesture}  & 85.94  &   80.00 &83.13 & 82.50 &  85.31\\
    \midrule
    \textbf{Average}  & 72.44 & 68.62   & 70.01  &  69.60 & 71.82 \\
    \bottomrule
\end{tabular}
}
\end{table}

\textbf{Results.} The results from Table \ref{ablation_table} indicate the importance of every main component of the overall design of CAB. First, while MOHA is disabled, the pure-CAB-Transformer obtains poor performance, as opposed to the other three ablation variants. Second, despite being in its simplified version, lagged cross-correlation is crucial and demonstrates significant improvement from pure-Transformer+CAB to  static-Transformer+CAB. The decrease in performance of $\lambda$-Transformer+CAB from  static-Transformer+CAB demonstrates that for low-dimensional MTS data, leanrable $\lambda$ is unnecessary. Nevertheless, for high-dimensional data, such as FaceDetection, learnable $\lambda$ results in efficiency gain. Finally, from the good performance of $\beta$-Transformer+CAB, we conclude that the score aggregation step with learnable $\beta$ is important in the pipeline of CAB. All in all, for all the ablation versions, the drops in accuracies are insignificant, thereby  showing the robustness of our model. 
 Furthermore, the two most crucial components of the CAB that account for the most significant efficiency boost are the  lagged cross-correlation filtering (as shown in static-Transformer+CAB versus pure-CAB-Transformer) and the learnable $\beta$ for balancing between instantaneous and lagged cross-correlation in Equation \ref{letstopk3} (as shown in $\beta$-Transformer+CAB versus static-Transformer+CAB).

\section{Additional Experimental Results for Long-term Forecasting}
\label{appen_forecast}

The decoder architecture of many prevalent Transformers (e.g. \citep{NIPS2017_3f5ee243, liu2022nonstationary}) is comprised of a masked multi-head attention block and a usual multi-head attention block (i.e. without masking). 
In its current form,  CAB has not been designed to be fully integrated into decoder architecture of Transformers yet, since it lacks the suitable masking mechanism. 
Nevertheless, we still consider the naive design of decoder architecture that still maintains the masked multi-head attention block of the base Transformer, yet (for the non-masked block) deploys mixture-of-head attention block combining the base temporal attention with CAB. We then test the effectiveness of the above decoder architecture integrated with CAB in MTS long-term forecasting, and believe that with proper masking mechanism for CAB in the future work, the performance increase can be further improved. 
Specifically, we augment Nonstationary Transformer \citep{liu2022nonstationary}, well-known for its competitive performance in long-term forecasting, with CAB, and experiment Nonstationary+CAB on the two common datasets ETTh \citep{zhou2021informer} (using ETTh2 as a representative), Weather \citep{weather_data} and Exchange \citep{lai2018modeling}. We follow the experimental settings of \citep{wu2022autoformer, wu2023timesnet} where the past sequence length is set to 96, and the prediction length is one of $\{ 96, 192, 336, 720\}$. We compare the empirical performance with the latest and state-of-the-art Transformer-models including Nonstationary Transformer \citep{liu2022nonstationary}, Fedformer \citep{fedformer1} and Autoformer \citep{wu2022autoformer}.

\begin{table}[H]
\caption{Long-term forecasting task on ETTh2, Weather and Exchange.}
\label{forecasting_table}
\resizebox{\linewidth}{!}{
\begin{tabular}{S|S|SS|SS|SS|SS} \toprule
\centering
    {Datasets} & {\makecell{Prediction \\Length}} &\multicolumn{2}{c}{\makecell{Nonstationary \\ \citep{liu2022nonstationary}}} & \multicolumn{2}{c}{\makecell{Nonstationary+CAB \\ \textbf{(Ours)}}}  &\multicolumn{2}{c}{\makecell{Fedformer  \\ \citep{fedformer1}}}   &\multicolumn{2}{c}{\makecell{Autoformer  \\ \citep{wu2022autoformer}}}  \\ 
    \midrule
    {} & {} & {MSE}&{MAE}& {MSE}&{ MAE} & {MSE}&{ MAE} & {MSE}&{ MAE}  \\
    \midrule
    \text{ETTh2}  & {96} & 0.476 & 0.458 &   0.376 & 0.407  & 0.358 & 0.397   &  0.346 & 0.388 \\
    \text{ETTh2}  & {192} & 0.512 & 0.493    &  0.513 & 0.476  &   0.429 & 0.439 &   0.456 & 0.452\\
    \text{ETTh2}  & {336} &  0.552 & 0.551  &  0.522 & 0.486  &  0.496 & 0.487 & 0.482 & 0.486  \\
    \text{ETTh2}  & {720} &  0.562  & 0.560   &  0.549 & 0.508 &  0.463 & 0.474&  0.515 & 0.511 \\
    \midrule
    \textbf{Average}  & {} & 0.526 & 0.516   &  0.490 & 0.469  & 0.437  &  0.449  &    0.450 & 0.459 \\
    \midrule
    \text{Weather} & {96 }  & 0.173 & 0.223    &   0.189 & 0.241  &  0.217 & 0.296  &  0.266 & 0.336 \\
    \text{Weather} & {192}  & 0.245 & 0.285    &    0.242 & 0.285 &   0.276 & 0.336 & 0.307 & 0.367\\
    \text{Weather} & {336 }  &0.321  & 0.338   &   0.307 & 0.333  &  0.339 & 0.380   &  0.359 & 0.395  \\
    \text{Weather} & {720 }  &0.414 & 0.410  &     0.379 & 0.382& 0.403 & 0.428   &  0.419 & 0.428 \\
    \midrule
    \textbf{Average}  & {} & 0.288 &0.314   &  0.279 & 0.310& 0.309 & 0.360   &  0.338 & 0.382\\
    \midrule
    \text{Exchange} & {96 }  &  0.111 & 0.237  &  0.123 & 0.249     &  0.148 & 0.278  & 0.197 & 0.323\\
    \text{Exchange} & {192}  &   0.219 & 0.335 &    0.224 & 0.340     &  0.271 & 0.380   & 0.300 & 0.369 \\
    \text{Exchange} & {336 }  & 0.421 & 0.476 &   0.327 & 0.416     & 0.460 & 0.500   & 0.509 & 0.524\\
    \text{Exchange} & {720 }  & 1.092 & 0.769 &  0.983 & 0.757     &   1.195 & 0.841  &  1.447 & 0.941\\
    \midrule
    \textbf{Average}  & {} &  0.461 & 0.454 &  0.414&  0.440  &  0.519 & 0.500    & 0.613 & 0.539  \\
    \bottomrule
\end{tabular}
}
\end{table}

\textbf{Results.} As shown in Table \ref{forecasting_table}, CAB, when integrated with Nonstationary Transformer, consistently improves the performance of the base model on all of the considered datasets spanning different disciplines. Moreover, Nonstationary+CAB even achieves the best performance among the baselines on the Weather and Exchange datasets.
This demonstrates the potential of CAB and the mixture-of-head attention design even in MTS predictive tasks.

\section{Run-time Analysis}
\label{appen_runtime}

We further provide performance measurement of the baselines for the imputation task on ETTh \citep{zhou2021informer} (using ETTh1 as a representative). In Table \ref{runtime_table}, we report the average run-time per iteration (s / iter) of all the methods tested in Section \ref{sec_impute}.

 \begin{table}[H]
 \centering
\caption{Run-time per iteration in (s / iter) for imputation task on ETTh1.}
\label{runtime_table}
\resizebox{\linewidth}{!}{
\begin{tabular}{S|S|S|S|S|S|S|S|S} \toprule
\centering
     {Series Length} &{\makecell{TimesNet \\ \citep{wu2023timesnet}}} & {\makecell{Nonstationary \\ \citep{liu2022nonstationary}}} & { \makecell{Nonstationary+CAB \\ \textbf{(Ours)} }} &{ \makecell{Transformer \\ \citep{NIPS2017_3f5ee243}} } &{\makecell{Transformer+CAB\\\textbf{(Ours)}}}  & { \makecell{FEDformer \\ \citep{fedformer1}} } & { \makecell{DLinear \\ \citep{zeng2022transformers}} } & { \makecell{Autoformer \\ \citep{wu2022autoformer}} } \\ 
    \midrule
     {384} &  0.024 &  0.046   & 0.069  & 0.024  & 0.067  &   0.807 & 0.006  & 0.070   \\
     {768} &  0.040 &  0.118  & 0.121   & 0.082  & 0.103  &   1.055 &  0.006 &  0.071  \\
     {1536} & 0.045  & 0.467   &  0.542 &  0.104 &  0.278  &  1.482  &   0.007 &   0.129 \\
    \bottomrule
\end{tabular}
}
\end{table}

\textbf{Results.} As shown in  Table \ref{runtime_table}, the CAB incurs only minimal overhead to the base Transformers, especially for the Nonstationary+CAB baseline which achieves the state-of-the-art results for imputation. We note that the two baselines TimesNet and DLinear with superior run-time performance are non-Transformer models and sub-optimal in their achievable error performance. 

%% file: CorrAttention_arxiv.bbl
\begin{thebibliography}{60}
\providecommand{\natexlab}[1]{#1}
\providecommand{\url}[1]{\texttt{#1}}
\expandafter\ifx\csname urlstyle\endcsname\relax
  \providecommand{\doi}[1]{doi: #1}\else
  \providecommand{\doi}{doi: \begingroup \urlstyle{rm}\Url}\fi

\bibitem[Abdulaal et~al.(2021)Abdulaal, Liu, and Lancewicki]{PSM1}
Ahmed Abdulaal, Zhuanghua Liu, and Tomer Lancewicki.
\newblock Practical approach to asynchronous multivariate time series anomaly detection and localization.
\newblock In \emph{Proceedings of the 27th ACM SIGKDD Conference on Knowledge Discovery \& Data Mining}, KDD '21, page 2485–2494, New York, NY, USA, 2021. Association for Computing Machinery.
\newblock ISBN 9781450383325.
\newblock \doi{10.1145/3447548.3467174}.
\newblock URL \url{https://doi.org/10.1145/3447548.3467174}.

\bibitem[Bagnall et~al.(2018)Bagnall, Dau, Lines, Flynn, Large, Bostrom, Southam, and Keogh]{bagnall2018uea}
Anthony Bagnall, Hoang~Anh Dau, Jason Lines, Michael Flynn, James Large, Aaron Bostrom, Paul Southam, and Eamonn Keogh.
\newblock The uea multivariate time series classification archive, 2018, 2018.

\bibitem[Blázquez-García et~al.(2020)Blázquez-García, Conde, Mori, and Lozano]{blázquezgarcía2020review}
Ane Blázquez-García, Angel Conde, Usue Mori, and Jose~A. Lozano.
\newblock A review on outlier/anomaly detection in time series data, 2020.

\bibitem[Cao et~al.(2020)Cao, Wang, Duan, Zhang, Zhu, Huang, Tong, Xu, Bai, Tong, and Zhang]{spectral_temporal1}
Defu Cao, Yujing Wang, Juanyong Duan, Ce~Zhang, Xia Zhu, Conguri Huang, Yunhai Tong, Bixiong Xu, Jing Bai, Jie Tong, and Qi~Zhang.
\newblock Spectral temporal graph neural network for multivariate time-series forecasting.
\newblock In \emph{Proceedings of the 34th International Conference on Neural Information Processing Systems}, NIPS'20, Red Hook, NY, USA, 2020. Curran Associates Inc.
\newblock ISBN 9781713829546.

\bibitem[Challu et~al.(2022)Challu, Olivares, Oreshkin, Garza, Mergenthaler-Canseco, and Dubrawski]{challu2022nhits}
Cristian Challu, Kin~G. Olivares, Boris~N. Oreshkin, Federico Garza, Max Mergenthaler-Canseco, and Artur Dubrawski.
\newblock N-hits: Neural hierarchical interpolation for time series forecasting, 2022.

\bibitem[Chandereng and Gitter(2020)]{Chandereng2020-zz}
Thevaa Chandereng and Anthony Gitter.
\newblock Lag penalized weighted correlation for time series clustering.
\newblock \emph{BMC Bioinformatics}, 21\penalty0 (1):\penalty0 21, August 2020.

\bibitem[Chatfield(2004)]{chatfield2004timeseries}
Chris Chatfield.
\newblock \emph{The analysis of time series: an introduction}.
\newblock CRC Press, Florida, US, 6th edition, 2004.

\bibitem[Cirstea et~al.(2018)Cirstea, Micu, Muresan, Guo, and Yang]{cirstea2018correlated}
Razvan-Gabriel Cirstea, Darius-Valer Micu, Gabriel-Marcel Muresan, Chenjuan Guo, and Bin Yang.
\newblock Correlated time series forecasting using deep neural networks: A summary of results, 2018.

\bibitem[Cirstea et~al.(2021)Cirstea, Kieu, Guo, Yang, and Pan]{correlatedMTS6}
Razvan-Gabriel Cirstea, Tung Kieu, Chenjuan Guo, Bin Yang, and Sinno~Jialin Pan.
\newblock Enhancenet: Plugin neural networks for enhancing correlated time series forecasting.
\newblock In \emph{2021 IEEE 37th International Conference on Data Engineering (ICDE)}, pages 1739--1750, 2021.
\newblock \doi{10.1109/ICDE51399.2021.00153}.

\bibitem[Contreras-Reyes and Idrovo-Aguirre(2020)]{CONTRERASREYES2020125109}
Javier~E. Contreras-Reyes and Byron~J. Idrovo-Aguirre.
\newblock Backcasting and forecasting time series using detrended cross-correlation analysis.
\newblock \emph{Physica A: Statistical Mechanics and its Applications}, 560:\penalty0 125109, 2020.
\newblock ISSN 0378-4371.
\newblock \doi{https://doi.org/10.1016/j.physa.2020.125109}.
\newblock URL \url{https://www.sciencedirect.com/science/article/pii/S0378437120305768}.

\bibitem[Du et~al.(2023{\natexlab{a}})Du, C{\^{o}}t{\'{e}}, and Liu]{Du_2023}
Wenjie Du, David C{\^{o}}t{\'{e}}, and Yan Liu.
\newblock {SAITS}: Self-attention-based imputation for time series.
\newblock \emph{Expert Systems with Applications}, 219:\penalty0 119619, jun 2023{\natexlab{a}}.
\newblock \doi{10.1016/j.eswa.2023.119619}.
\newblock URL \url{https://doi.org/10.1016\%2Fj.eswa.2023.119619}.

\bibitem[Du et~al.(2023{\natexlab{b}})Du, CÃŽtÃ©, and Liu]{DU2023119619}
Wenjie Du, David CÃŽtÃ©, and Yan Liu.
\newblock Saits: Self-attention-based imputation for time series.
\newblock \emph{Expert Systems with Applications}, 219:\penalty0 119619, 2023{\natexlab{b}}.
\newblock ISSN 0957-4174.
\newblock \doi{https://doi.org/10.1016/j.eswa.2023.119619}.
\newblock URL \url{https://www.sciencedirect.com/science/article/pii/S0957417423001203}.

\bibitem[El-Nouby et~al.(2021)El-Nouby, Touvron, Caron, Bojanowski, Douze, Joulin, Laptev, Neverova, Synnaeve, Verbeek, and Jegou]{elnouby2021xcit}
Alaaeldin El-Nouby, Hugo Touvron, Mathilde Caron, Piotr Bojanowski, Matthijs Douze, Armand Joulin, Ivan Laptev, Natalia Neverova, Gabriel Synnaeve, Jakob Verbeek, and Hervé Jegou.
\newblock Xcit: Cross-covariance image transformers, 2021.

\bibitem[Esling and Agon(2012)]{lit2}
Philippe Esling and Carlos Agon.
\newblock Time-series data mining.
\newblock \emph{ACM Comput. Surv.}, 45\penalty0 (1), dec 2012.
\newblock ISSN 0360-0300.
\newblock \doi{10.1145/2379776.2379788}.
\newblock URL \url{https://doi.org/10.1145/2379776.2379788}.

\bibitem[Fawaz et~al.(2019)Fawaz, Forestier, Weber, Idoumghar, and Muller]{Ismail_Fawaz_2019}
Hassan~Ismail Fawaz, Germain Forestier, Jonathan Weber, Lhassane Idoumghar, and Pierre-Alain Muller.
\newblock Deep learning for time series classification: a review.
\newblock \emph{Data Mining and Knowledge Discovery}, 33\penalty0 (4):\penalty0 917--963, mar 2019.
\newblock \doi{10.1007/s10618-019-00619-1}.
\newblock URL \url{https://doi.org/10.1007%2Fs10618-019-00619-1}.

\bibitem[Franceschi et~al.(2020)Franceschi, Dieuleveut, and Jaggi]{franceschi2020unsupervised}
Jean-Yves Franceschi, Aymeric Dieuleveut, and Martin Jaggi.
\newblock Unsupervised scalable representation learning for multivariate time series, 2020.

\bibitem[Gu et~al.(2022)Gu, Goel, and Ré]{gu2022efficiently}
Albert Gu, Karan Goel, and Christopher Ré.
\newblock Efficiently modeling long sequences with structured state spaces, 2022.

\bibitem[Hochreiter and Schmidhuber(1997)]{lstm1}
Sepp Hochreiter and J\"{u}rgen Schmidhuber.
\newblock Long short-term memory.
\newblock \emph{Neural Comput.}, 9\penalty0 (8):\penalty0 1735–1780, nov 1997.
\newblock ISSN 0899-7667.
\newblock \doi{10.1162/neco.1997.9.8.1735}.
\newblock URL \url{https://doi.org/10.1162/neco.1997.9.8.1735}.

\bibitem[Hundman et~al.(2018{\natexlab{a}})Hundman, Constantinou, Laporte, Colwell, and Soderstrom]{MSL1}
Kyle Hundman, Valentino Constantinou, Christopher Laporte, Ian Colwell, and Tom Soderstrom.
\newblock Detecting spacecraft anomalies using lstms and nonparametric dynamic thresholding.
\newblock In \emph{Proceedings of the 24th ACM SIGKDD International Conference on Knowledge Discovery \& Data Mining}, KDD '18, page 387–395, New York, NY, USA, 2018{\natexlab{a}}. Association for Computing Machinery.
\newblock ISBN 9781450355520.
\newblock \doi{10.1145/3219819.3219845}.
\newblock URL \url{https://doi.org/10.1145/3219819.3219845}.

\bibitem[Hundman et~al.(2018{\natexlab{b}})Hundman, Constantinou, Laporte, Colwell, and Soderstrom]{spacecraft1}
Kyle Hundman, Valentino Constantinou, Christopher Laporte, Ian Colwell, and Tom Soderstrom.
\newblock Detecting spacecraft anomalies using lstms and nonparametric dynamic thresholding.
\newblock In \emph{Proceedings of the 24th ACM SIGKDD International Conference on Knowledge Discovery \& Data Mining}, KDD '18, page 387–395, New York, NY, USA, 2018{\natexlab{b}}. Association for Computing Machinery.
\newblock ISBN 9781450355520.
\newblock \doi{10.1145/3219819.3219845}.
\newblock URL \url{https://doi.org/10.1145/3219819.3219845}.

\bibitem[John and Ferbinteanu(2021)]{John2021-mr}
Majnu John and Janina Ferbinteanu.
\newblock Detecting time lag between a pair of time series using visibility graph algorithm.
\newblock \emph{Commun Stat Case Stud Data Anal Appl}, 7\penalty0 (3):\penalty0 315--343, February 2021.

\bibitem[Kampouraki et~al.(2009)Kampouraki, Manis, and Nikou]{heartbeat1}
Argyro Kampouraki, George Manis, and Christophoros Nikou.
\newblock Heartbeat time series classification with support vector machines.
\newblock \emph{IEEE Transactions on Information Technology in Biomedicine}, 13\penalty0 (4):\penalty0 512--518, 2009.
\newblock \doi{10.1109/TITB.2008.2003323}.

\bibitem[Kieu et~al.(2018)Kieu, Yang, and Jensen]{correlatedMTS4}
Tung Kieu, Bin Yang, and Christian~S. Jensen.
\newblock Outlier detection for multidimensional time series using deep neural networks.
\newblock In \emph{2018 19th IEEE International Conference on Mobile Data Management (MDM)}, pages 125--134, 2018.
\newblock \doi{10.1109/MDM.2018.00029}.

\bibitem[Kristoufek(2014)]{KRISTOUFEK2014291}
Ladislav Kristoufek.
\newblock Measuring correlations between non-stationary series with dcca coefficient.
\newblock \emph{Physica A: Statistical Mechanics and its Applications}, 402:\penalty0 291--298, 2014.
\newblock ISSN 0378-4371.
\newblock \doi{https://doi.org/10.1016/j.physa.2014.01.058}.
\newblock URL \url{https://www.sciencedirect.com/science/article/pii/S037843711400079X}.

\bibitem[Lahiri(2016)]{cross_corr_theorem}
Avijit Lahiri.
\newblock Chapter 6 - fourier optics.
\newblock In Avijit Lahiri, editor, \emph{Basic Optics}, pages 539--603. Elsevier, Amsterdam, 2016.
\newblock ISBN 978-0-12-805357-7.
\newblock \doi{https://doi.org/10.1016/B978-0-12-805357-7.00006-X}.
\newblock URL \url{https://www.sciencedirect.com/science/article/pii/B978012805357700006X}.

\bibitem[Lai et~al.(2018)Lai, Chang, Yang, and Liu]{lai2018modeling}
Guokun Lai, Wei-Cheng Chang, Yiming Yang, and Hanxiao Liu.
\newblock Modeling long- and short-term temporal patterns with deep neural networks, 2018.

\bibitem[Lea et~al.(2016)Lea, Flynn, Vidal, Reiter, and Hager]{lea2016temporal}
Colin Lea, Michael~D. Flynn, Rene Vidal, Austin Reiter, and Gregory~D. Hager.
\newblock Temporal convolutional networks for action segmentation and detection, 2016.

\bibitem[Li et~al.(2021{\natexlab{a}})Li, Ling, and Yu]{correlatedMTS2}
Bao-Gen Li, Dian-Yi Ling, and Zu-Guo Yu.
\newblock Multifractal temporally weighted detrended partial cross-correlation analysis of two non-stationary time series affected by common external factors.
\newblock \emph{Physica A: Statistical Mechanics and its Applications}, 573:\penalty0 125920, 2021{\natexlab{a}}.
\newblock ISSN 0378-4371.
\newblock \doi{https://doi.org/10.1016/j.physa.2021.125920}.
\newblock URL \url{https://www.sciencedirect.com/science/article/pii/S0378437121001928}.

\bibitem[Li et~al.(2021{\natexlab{b}})Li, Zhang, Yan, Jin, Zhang, Duan, and Tian]{Li2021SynergeticLO}
Longyuan Li, Jihai Zhang, Junchi Yan, Yaohui Jin, Yunhao Zhang, Yanjie Duan, and Guangjian Tian.
\newblock Synergetic learning of heterogeneous temporal sequences for multi-horizon probabilistic forecasting.
\newblock In \emph{AAAI Conference on Artificial Intelligence}, 2021{\natexlab{b}}.
\newblock URL \url{https://api.semanticscholar.org/CorpusID:231741193}.

\bibitem[Li et~al.(2019)Li, Jin, Xuan, Zhou, Chen, Wang, and Yan]{enhance_memory1}
Shiyang Li, Xiaoyong Jin, Yao Xuan, Xiyou Zhou, Wenhu Chen, Yu-Xiang Wang, and Xifeng Yan.
\newblock \emph{Enhancing the Locality and Breaking the Memory Bottleneck of Transformer on Time Series Forecasting}.
\newblock Curran Associates Inc., Red Hook, NY, USA, 2019.

\bibitem[Lim and Zohren(2021)]{Lim_2021}
Bryan Lim and Stefan Zohren.
\newblock Time-series forecasting with deep learning: a survey.
\newblock \emph{Philosophical Transactions of the Royal Society A: Mathematical, Physical and Engineering Sciences}, 379\penalty0 (2194):\penalty0 20200209, feb 2021.
\newblock \doi{10.1098/rsta.2020.0209}.
\newblock URL \url{https://doi.org/10.1098%2Frsta.2020.0209}.

\bibitem[Liu et~al.(2022)Liu, Wu, Wang, and Long]{liu2022nonstationary}
Yong Liu, Haixu Wu, Jianmin Wang, and Mingsheng Long.
\newblock Non-stationary transformers: Exploring the stationarity in time series forecasting.
\newblock In Alice~H. Oh, Alekh Agarwal, Danielle Belgrave, and Kyunghyun Cho, editors, \emph{Advances in Neural Information Processing Systems}, 2022.
\newblock URL \url{https://openreview.net/forum?id=ucNDIDRNjjv}.

\bibitem[López et~al.(2021)López, Hernández, Urrutia, López-Cortés, Araya, and Morales-Salinas]{LOPEZ2021104794}
J.L. López, S.~Hernández, A.~Urrutia, X.A. López-Cortés, H.~Araya, and L.~Morales-Salinas.
\newblock Effect of missing data on short time series and their application in the characterization of surface temperature by detrended fluctuation analysis.
\newblock \emph{Computers \& Geosciences}, 153:\penalty0 104794, 2021.
\newblock ISSN 0098-3004.
\newblock \doi{https://doi.org/10.1016/j.cageo.2021.104794}.
\newblock URL \url{https://www.sciencedirect.com/science/article/pii/S0098300421000960}.

\bibitem[Mathur and Tippenhauer(2016)]{SWAT1}
Aditya~P. Mathur and Nils~Ole Tippenhauer.
\newblock Swat: a water treatment testbed for research and training on ics security.
\newblock In \emph{2016 International Workshop on Cyber-physical Systems for Smart Water Networks (CySWater)}, pages 31--36, 2016.
\newblock \doi{10.1109/CySWater.2016.7469060}.

\bibitem[Oreshkin et~al.(2020)Oreshkin, Carpov, Chapados, and Bengio]{oreshkin2020nbeats}
Boris~N. Oreshkin, Dmitri Carpov, Nicolas Chapados, and Yoshua Bengio.
\newblock N-beats: Neural basis expansion analysis for interpretable time series forecasting, 2020.

\bibitem[Papoulis(1965)]{Papoulis1965ProbabilityRV}
Athanasios Papoulis.
\newblock Probability, random variables and stochastic processes.
\newblock 1965.
\newblock URL \url{https://api.semanticscholar.org/CorpusID:118245370}.

\bibitem[Probst et~al.(2012)Probst, Stelzenmüller, and Fock]{lagged_cross1}
Wolfgang~Nikolaus Probst, Vanessa Stelzenmüller, and Heino~Ove Fock.
\newblock {Using cross-correlations to assess the relationship between time-lagged pressure and state indicators: an exemplary analysis of North Sea fish population indicators}.
\newblock \emph{ICES Journal of Marine Science}, 69\penalty0 (4):\penalty0 670--681, 02 2012.
\newblock ISSN 1054-3139.
\newblock \doi{10.1093/icesjms/fss015}.
\newblock URL \url{https://doi.org/10.1093/icesjms/fss015}.

\bibitem[Shen(2015)]{SHEN2015680}
Chenhua Shen.
\newblock Analysis of detrended time-lagged cross-correlation between two nonstationary time series.
\newblock \emph{Physics Letters A}, 379\penalty0 (7):\penalty0 680--687, 2015.
\newblock ISSN 0375-9601.
\newblock \doi{https://doi.org/10.1016/j.physleta.2014.12.036}.
\newblock URL \url{https://www.sciencedirect.com/science/article/pii/S0375960114012766}.

\bibitem[Shen et~al.(2020)Shen, Li, and Kwok]{NEURIPS2020_97e401a0}
Lifeng Shen, Zhuocong Li, and James Kwok.
\newblock Timeseries anomaly detection using temporal hierarchical one-class network.
\newblock In H.~Larochelle, M.~Ranzato, R.~Hadsell, M.F. Balcan, and H.~Lin, editors, \emph{Advances in Neural Information Processing Systems}, volume~33, pages 13016--13026. Curran Associates, Inc., 2020.
\newblock URL \url{https://proceedings.neurips.cc/paper_files/paper/2020/file/97e401a02082021fd24957f852e0e475-Paper.pdf}.

\bibitem[Su et~al.(2019)Su, Zhao, Niu, Liu, Sun, and Pei]{SMD1}
Ya~Su, Youjian Zhao, Chenhao Niu, Rong Liu, Wei Sun, and Dan Pei.
\newblock Robust anomaly detection for multivariate time series through stochastic recurrent neural network.
\newblock In \emph{Proceedings of the 25th ACM SIGKDD International Conference on Knowledge Discovery \& Data Mining}, KDD '19, page 2828–2837, New York, NY, USA, 2019. Association for Computing Machinery.
\newblock ISBN 9781450362016.
\newblock \doi{10.1145/3292500.3330672}.
\newblock URL \url{https://doi.org/10.1145/3292500.3330672}.

\bibitem[Trindade(2015)]{misc_electricityloaddiagrams20112014_321}
Artur Trindade.
\newblock {ElectricityLoadDiagrams20112014}.
\newblock UCI Machine Learning Repository, 2015.
\newblock {DOI}: https://doi.org/10.24432/C58C86.

\bibitem[Vaswani et~al.(2017)Vaswani, Shazeer, Parmar, Uszkoreit, Jones, Gomez, Kaiser, and Polosukhin]{NIPS2017_3f5ee243}
Ashish Vaswani, Noam Shazeer, Niki Parmar, Jakob Uszkoreit, Llion Jones, Aidan~N Gomez, \L~ukasz Kaiser, and Illia Polosukhin.
\newblock Attention is all you need.
\newblock In I.~Guyon, U.~Von Luxburg, S.~Bengio, H.~Wallach, R.~Fergus, S.~Vishwanathan, and R.~Garnett, editors, \emph{Advances in Neural Information Processing Systems}, volume~30. Curran Associates, Inc., 2017.
\newblock URL \url{https://proceedings.neurips.cc/paper_files/paper/2017/file/3f5ee243547dee91fbd053c1c4a845aa-Paper.pdf}.

\bibitem[Wen et~al.(2022)Wen, Yang, Zhou, and Sun]{lit1}
Qingsong Wen, Linxiao Yang, Tian Zhou, and Liang Sun.
\newblock Robust time series analysis and applications: An industrial perspective.
\newblock In \emph{Proceedings of the 28th ACM SIGKDD Conference on Knowledge Discovery and Data Mining}, KDD '22, page 4836–4837, New York, NY, USA, 2022. Association for Computing Machinery.
\newblock ISBN 9781450393850.
\newblock \doi{10.1145/3534678.3542612}.
\newblock URL \url{https://doi.org/10.1145/3534678.3542612}.

\bibitem[Wen et~al.(2023)Wen, Zhou, Zhang, Chen, Ma, Yan, and Sun]{wen2023transformers}
Qingsong Wen, Tian Zhou, Chaoli Zhang, Weiqi Chen, Ziqing Ma, Junchi Yan, and Liang Sun.
\newblock Transformers in time series: A survey, 2023.

\bibitem[Wetterstation()]{weather_data}
Wetterstation.
\newblock Weather.
\newblock URL \url{https://www.bgc-jena.mpg.de/wetter/.}

\bibitem[Wu et~al.(2022{\natexlab{a}})Wu, Wu, Xu, Wang, and Long]{Wu2022FlowformerLT}
Haixu Wu, Jialong Wu, Jiehui Xu, Jianmin Wang, and Mingsheng Long.
\newblock Flowformer: Linearizing transformers with conservation flows.
\newblock In \emph{International Conference on Machine Learning}, 2022{\natexlab{a}}.
\newblock URL \url{https://api.semanticscholar.org/CorpusID:246822433}.

\bibitem[Wu et~al.(2022{\natexlab{b}})Wu, Xu, Wang, and Long]{wu2022autoformer}
Haixu Wu, Jiehui Xu, Jianmin Wang, and Mingsheng Long.
\newblock Autoformer: Decomposition transformers with auto-correlation for long-term series forecasting, 2022{\natexlab{b}}.

\bibitem[Wu et~al.(2023)Wu, Hu, Liu, Zhou, Wang, and Long]{wu2023timesnet}
Haixu Wu, Tengge Hu, Yong Liu, Hang Zhou, Jianmin Wang, and Mingsheng Long.
\newblock Timesnet: Temporal 2d-variation modeling for general time series analysis.
\newblock In \emph{The Eleventh International Conference on Learning Representations}, 2023.
\newblock URL \url{https://openreview.net/forum?id=ju_Uqw384Oq}.

\bibitem[Wu et~al.(2021)Wu, Zhang, Guo, He, Yang, and Jensen]{correlatedMTS1}
Xinle Wu, Dalin Zhang, Chenjuan Guo, Chaoyang He, Bin Yang, and Christian~S. Jensen.
\newblock Autocts: Automated correlated time series forecasting.
\newblock \emph{Proc. VLDB Endow.}, 15\penalty0 (4):\penalty0 971–983, dec 2021.
\newblock ISSN 2150-8097.
\newblock \doi{10.14778/3503585.3503604}.
\newblock URL \url{https://doi.org/10.14778/3503585.3503604}.

\bibitem[Wu et~al.(2020)Wu, Pan, Long, Jiang, Chang, and Zhang]{mts_forecast_gnn}
Zonghan Wu, Shirui Pan, Guodong Long, Jing Jiang, Xiaojun Chang, and Chengqi Zhang.
\newblock Connecting the dots: Multivariate time series forecasting with graph neural networks.
\newblock In \emph{Proceedings of the 26th ACM SIGKDD International Conference on Knowledge Discovery \& Data Mining}, KDD '20, page 753–763, New York, NY, USA, 2020. Association for Computing Machinery.
\newblock ISBN 9781450379984.
\newblock \doi{10.1145/3394486.3403118}.
\newblock URL \url{https://doi.org/10.1145/3394486.3403118}.

\bibitem[Xu et~al.(2022)Xu, Wu, Wang, and Long]{xu2022anomaly}
Jiehui Xu, Haixu Wu, Jianmin Wang, and Mingsheng Long.
\newblock Anomaly transformer: Time series anomaly detection with association discrepancy, 2022.

\bibitem[Yang et~al.(2013{\natexlab{a}})Yang, Guo, and Jensen]{correlatedMTS3}
Bin Yang, Chenjuan Guo, and Christian~S. Jensen.
\newblock Travel cost inference from sparse, spatio temporally correlated time series using markov models.
\newblock \emph{Proc. VLDB Endow.}, 6\penalty0 (9):\penalty0 769–780, jul 2013{\natexlab{a}}.
\newblock ISSN 2150-8097.
\newblock \doi{10.14778/2536360.2536375}.
\newblock URL \url{https://doi.org/10.14778/2536360.2536375}.

\bibitem[Yang et~al.(2013{\natexlab{b}})Yang, Guo, and Jensen]{correlatedMTS5}
Bin Yang, Chenjuan Guo, and Christian~S. Jensen.
\newblock Travel cost inference from sparse, spatio temporally correlated time series using markov models.
\newblock \emph{Proc. VLDB Endow.}, 6\penalty0 (9):\penalty0 769–780, jul 2013{\natexlab{b}}.
\newblock ISSN 2150-8097.
\newblock \doi{10.14778/2536360.2536375}.
\newblock URL \url{https://doi.org/10.14778/2536360.2536375}.

\bibitem[Yu et~al.(2018)Yu, Yin, and Zhu]{Yu_2018}
Bing Yu, Haoteng Yin, and Zhanxing Zhu.
\newblock Spatio-temporal graph convolutional networks: A deep learning framework for traffic forecasting.
\newblock In \emph{Proceedings of the Twenty-Seventh International Joint Conference on Artificial Intelligence}. International Joint Conferences on Artificial Intelligence Organization, jul 2018.
\newblock \doi{10.24963/ijcai.2018/505}.
\newblock URL \url{https://doi.org/10.24963%2Fijcai.2018%2F505}.

\bibitem[Yuan et~al.(2016)Yuan, Xoplaki, Zhu, and Luterbacher]{Yuan2016}
Naiming Yuan, Elena Xoplaki, Congwen Zhu, and Juerg Luterbacher.
\newblock A novel way to detect correlations on multi-time scales, with temporal evolution and for multi-variables.
\newblock \emph{Scientific Reports}, 6\penalty0 (1):\penalty0 27707, Jun 2016.
\newblock ISSN 2045-2322.
\newblock \doi{10.1038/srep27707}.
\newblock URL \url{https://doi.org/10.1038/srep27707}.

\bibitem[Zeng et~al.(2022)Zeng, Chen, Zhang, and Xu]{zeng2022transformers}
Ailing Zeng, Muxi Chen, Lei Zhang, and Qiang Xu.
\newblock Are transformers effective for time series forecasting?, 2022.

\bibitem[Zerveas et~al.(2021)Zerveas, Jayaraman, Patel, Bhamidipaty, and Eickhoff]{classi_benchmark1}
George Zerveas, Srideepika Jayaraman, Dhaval Patel, Anuradha Bhamidipaty, and Carsten Eickhoff.
\newblock A transformer-based framework for multivariate time series representation learning.
\newblock In \emph{Proceedings of the 27th ACM SIGKDD Conference on Knowledge Discovery \& Data Mining}, KDD '21, page 2114–2124, New York, NY, USA, 2021. Association for Computing Machinery.
\newblock ISBN 9781450383325.
\newblock \doi{10.1145/3447548.3467401}.
\newblock URL \url{https://doi.org/10.1145/3447548.3467401}.

\bibitem[Zhang and Yan(2023)]{zhang2023crossformer}
Yunhao Zhang and Junchi Yan.
\newblock Crossformer: Transformer utilizing cross-dimension dependency for multivariate time series forecasting.
\newblock In \emph{The Eleventh International Conference on Learning Representations}, 2023.
\newblock URL \url{https://openreview.net/forum?id=vSVLM2j9eie}.

\bibitem[Zhou et~al.(2021)Zhou, Zhang, Peng, Zhang, Li, Xiong, and Zhang]{zhou2021informer}
Haoyi Zhou, Shanghang Zhang, Jieqi Peng, Shuai Zhang, Jianxin Li, Hui Xiong, and Wancai Zhang.
\newblock Informer: Beyond efficient transformer for long sequence time-series forecasting, 2021.

\bibitem[Zhou et~al.(2022)Zhou, Ma, Wen, Wang, Sun, and Jin]{fedformer1}
Tian Zhou, Ziqing Ma, Qingsong Wen, Xue Wang, Liang Sun, and Rong Jin.
\newblock {FED}former: Frequency enhanced decomposed transformer for long-term series forecasting.
\newblock In Kamalika Chaudhuri, Stefanie Jegelka, Le~Song, Csaba Szepesvari, Gang Niu, and Sivan Sabato, editors, \emph{Proceedings of the 39th International Conference on Machine Learning}, volume 162 of \emph{Proceedings of Machine Learning Research}, pages 27268--27286. PMLR, 17--23 Jul 2022.
\newblock URL \url{https://proceedings.mlr.press/v162/zhou22g.html}.

\end{thebibliography}
